\documentclass{article}

\usepackage{microtype}
\usepackage{graphicx}
\usepackage{caption}
\usepackage{subcaption}
\usepackage{booktabs} 
\usepackage{multirow}
\usepackage[utf8]{inputenc} 
\usepackage[T1]{fontenc}    

\usepackage{hyperref}
\hypersetup{
    colorlinks=true,
    linkcolor=BrickRed,
    citecolor=NavyBlue,
    urlcolor=NavyBlue
}
\usepackage{url}
\usepackage{placeins}
\usepackage{xurl}

\usepackage{pifont}   
\usepackage{xcolor}   
\usepackage[table]{xcolor}

\usepackage[preprint]{icml2026}
\usepackage{amsmath}
\usepackage{amssymb}
\usepackage{mathtools}
\usepackage{amsthm}
\usepackage{booktabs}

\usepackage{arydshln}
\usepackage[capitalize,noabbrev]{cleveref}

\theoremstyle{plain}

\theoremstyle{definition}

\theoremstyle{remark}

\newcommand{\SelvaMask}{\textsc{SelvaMask}~}
\newcommand{\eg}{\textit{e.g.}~}
\newcommand{\ie}{\textit{i.e.}~}

\usepackage[textsize=tiny]{todonotes}

\icmltitlerunning{SelvaMask}
\newcommand{\teaserfigure}{
  \begin{center}
    \centering
    \captionsetup{type=figure}
    
    \begin{center}
    \setlength{\tabcolsep}{1pt} 
    \begin{tabular}{cccc}
      \includegraphics[width=0.23\textwidth]{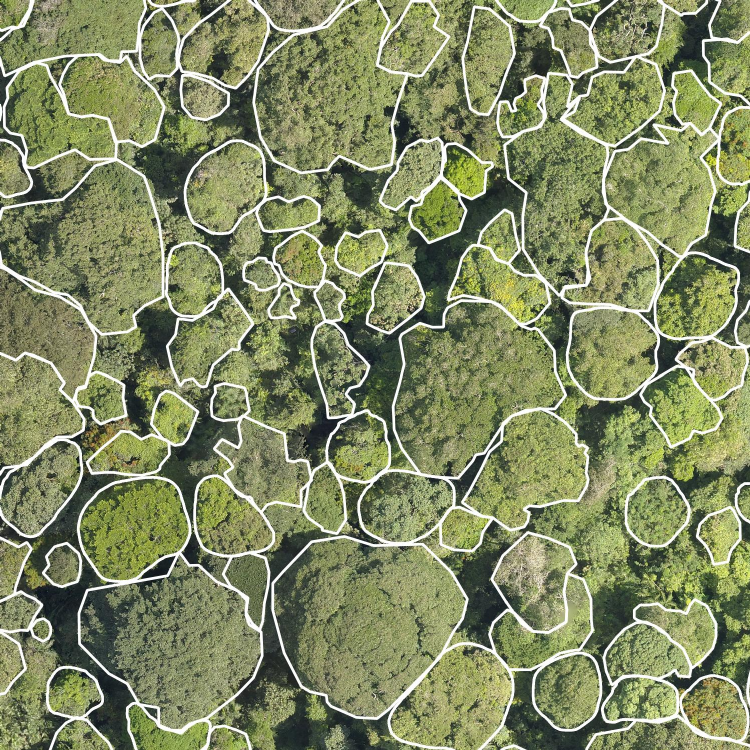} &
      \includegraphics[width=0.23\textwidth]{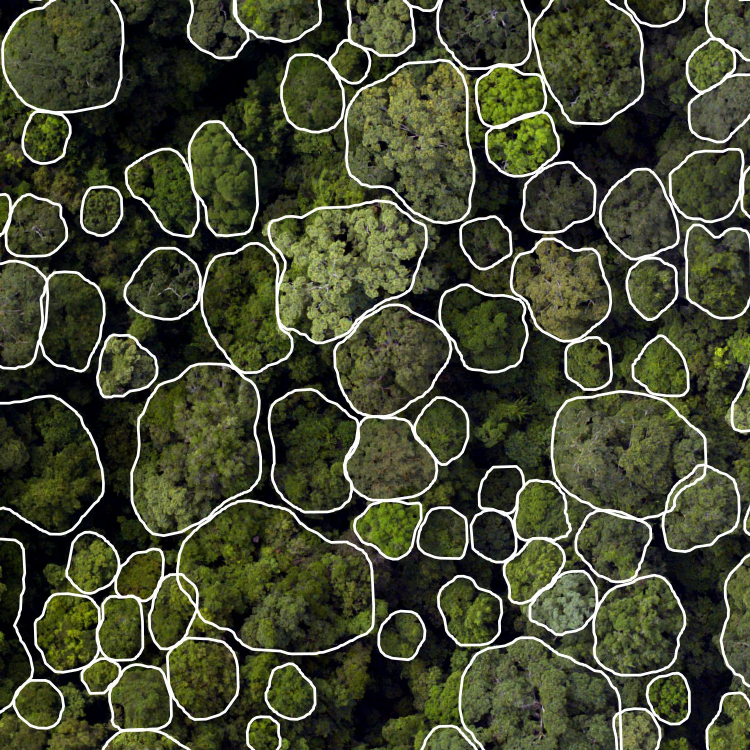} &
      \includegraphics[width=0.23\textwidth]{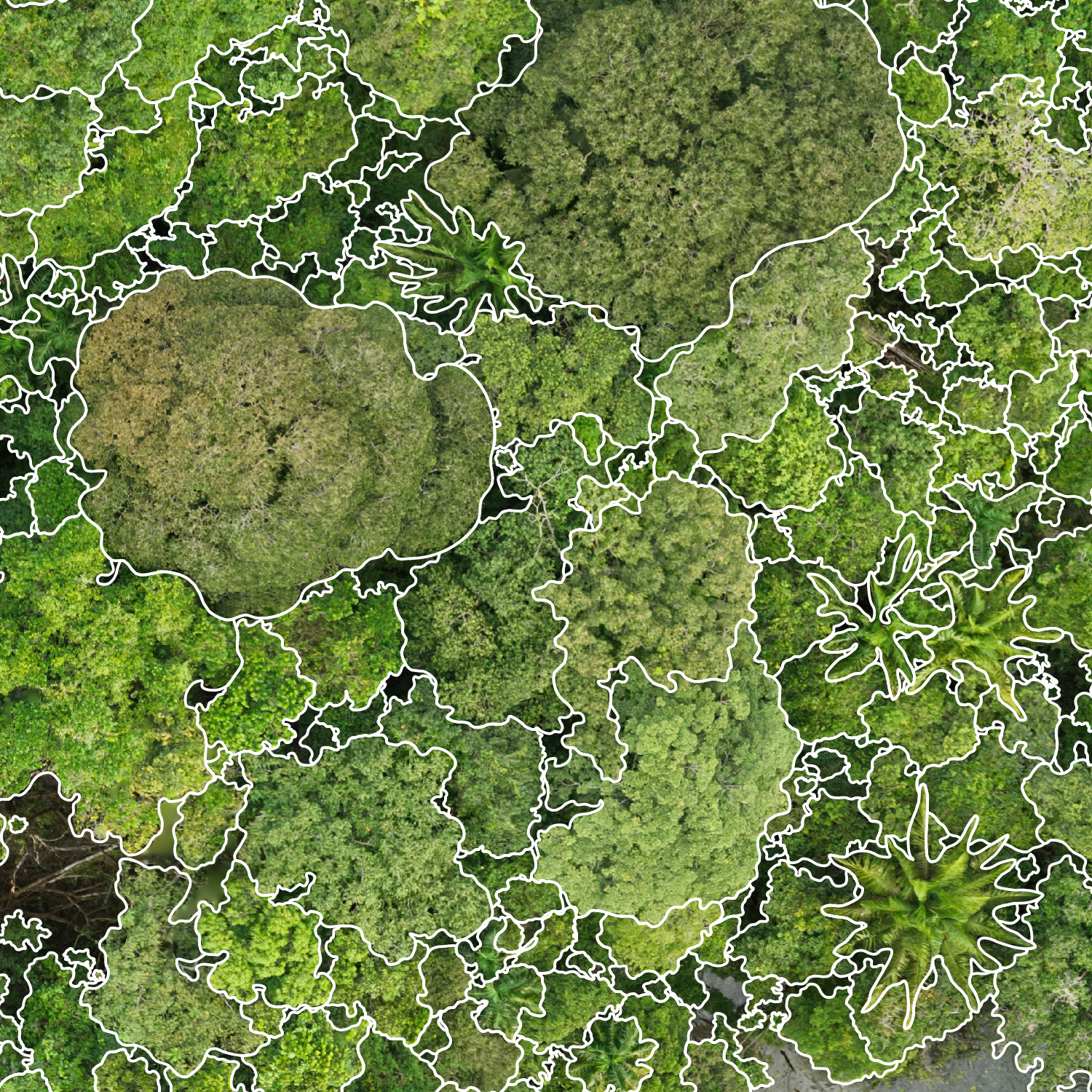} &
      \includegraphics[width=0.23\textwidth]{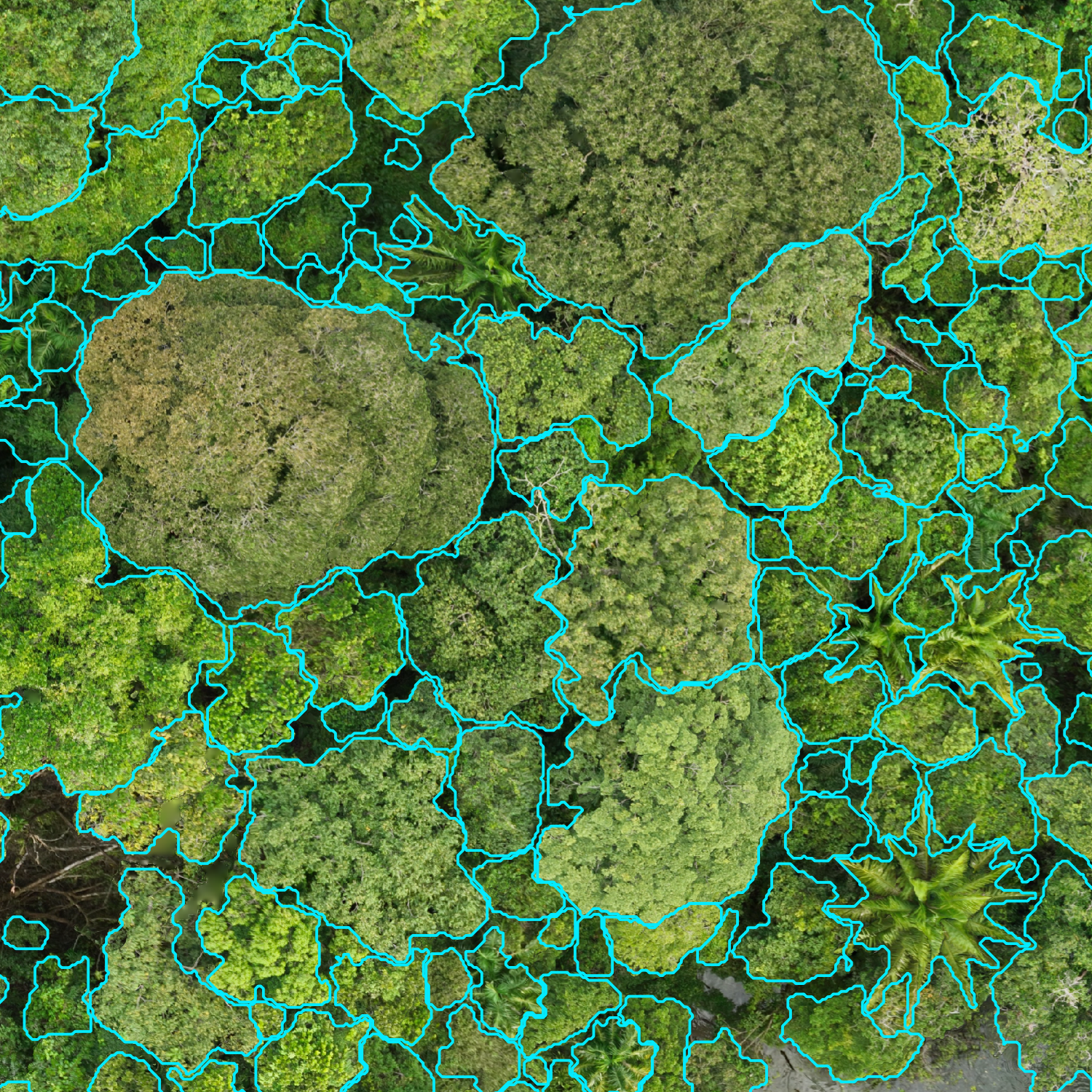} \\
      \footnotesize BCI50ha & \footnotesize Detectree2 & \footnotesize SelvaMask (ours) & \footnotesize Predictions (ours)
    \end{tabular}
    \end{center} 
    \caption{\textbf{Qualitative comparison.} Existing tropical forest datasets (BCI50ha, Detectree2) and our proposed \SelvaMask dataset, including the predictions of our modular pipeline for individual tree crown segmentation (see Sec.~\ref{sec:modular_pipeline}) on the \SelvaMask's test set.}
    \label{fig:qualitative_teaser}
  \end{center}
}
\newcommand{\fire}{\includegraphics[height=0.9em]{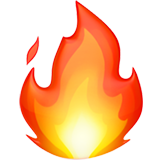}}
\newcommand{\ice}{\includegraphics[height=0.9em]{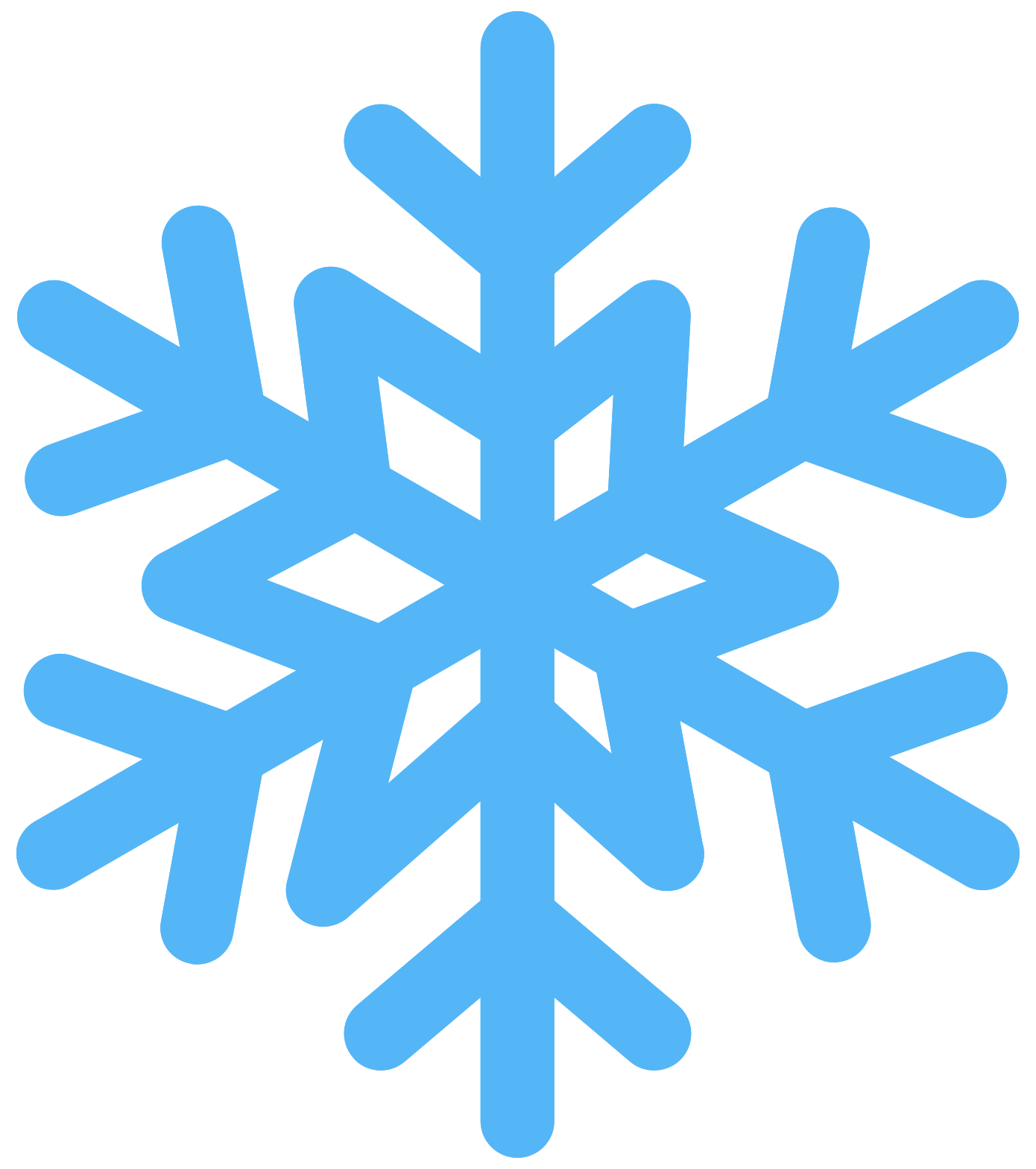}}
\newcommand{\cmark}{\textcolor{green!60!black}{\ding{51}}}
\newcommand{\xmark}{\textcolor{red}{\ding{55}}}

\begin{document}

\twocolumn[
  \icmltitle{\SelvaMask: Segmenting Trees in Tropical Forests and Beyond
  \vspace{0.5cm}
  }


  \icmlsetsymbol{equal}{*}

  \begin{icmlauthorlist}
    \icmlauthor{Simon-Olivier Duguay}{equal,udem,mila}
    \icmlauthor{Hugo Baudchon}{udem,mila}
    \icmlauthor{Etienne Laliberté}{udem,mila}
    \icmlauthor{Helene Muller-Landau}{smithsonian}
    \icmlauthor{Gonzalo Rivas-Torres}{quito}
    \icmlauthor{Arthur Ouaknine}{mcgill,mila}
  \end{icmlauthorlist}

  \icmlaffiliation{udem}{Université de Montréal}
  \icmlaffiliation{mcgill}{McGill University}
  \icmlaffiliation{mila}{Mila, Quebec AI Institute}
  \icmlaffiliation{smithsonian}{Smithsonian Tropical Research Institute}
  \icmlaffiliation{quito}{Universidad San Francisco de Quito}

  \icmlcorrespondingauthor{Simon-Olivier Duguay}{simon-olivier.duguay@umontreal.ca}

  \icmlkeywords{Machine Learning, ICML}

  \vskip 0.1in
  \teaserfigure

  \vskip 0.3in
]



\printAffiliationsAndNotice{}  

\begin{abstract}
Tropical forests harbor most of the planet’s tree biodiversity and are critical to global ecological balance.
Canopy trees in particular play a disproportionate role in carbon storage and functioning of these ecosystems.
Studying canopy trees at scale requires accurate delineation of individual tree crowns, typically performed using high-resolution aerial imagery. 
Despite advances in transformer-based models for individual tree crown segmentation, performance remains low in most forests, especially tropical ones.
To this end, we introduce \textsc{SelvaMask}, a new tropical dataset containing over 8\,800 manually delineated tree crowns across three Neotropical forest sites in Panama, Brazil, and Ecuador. \textsc{SelvaMask} features comprehensive annotations, including an inter-annotator agreement evaluation, capturing the dense structure of tropical forests and highlighting the difficulty of the task. Leveraging this benchmark, we propose a modular detection-segmentation pipeline that adapts vision foundation models (VFMs), using domain-specific detection-prompter. Our approach reaches state-of-the-art performance, outperforming both zero-shot generalist models and fully supervised end-to-end methods in dense tropical forests. 
We validate these gains on external tropical and temperate datasets, demonstrating that \SelvaMask serves as both a challenging benchmark and a key enabler for generalized forest monitoring. 
Our code\footnote{Benchmark, inference and training pipelines are integrated into \textit{CanopyRS} : \url{https://github.com/hugobaudchon/CanopyRS}} and dataset\footnote{\SelvaMask dataset: \url{https://huggingface.co/datasets/CanopyRS/SelvaMask}} will be released publicly.
\end{abstract}

\begin{table*}[t]
\centering
\footnotesize
\setlength{\tabcolsep}{4pt}
\renewcommand{\arraystretch}{1.08}
\begin{tabular}{@{}lcccccc@{}}
\toprule
\textbf{Dataset} & \textbf{Biome} & \textbf{Type} & \textbf{\#Crowns} & \textbf{GSD (cm/px)} & \textbf{Scope} \\
\midrule
Firoze et al. \cite{firozeTreeInstanceSegmentation2023} & temperate & natural & 6.5k & 2--5 & All Crowns\\
BAMFORESTS \cite{trolesBAMFORESTSBambergBenchmark2024} & temperate & natural & 27k & 1.6--1.8  & All Crowns\\
QuebecTrees \cite{cloutierInfluenceTemperateForest2024} & temperate & natural &  23k & 1.9 & Subset \\
Quebec Plantation \cite{lefebvreisabelle2024quebecplantations} & temperate & plantation & 19.6k & 0.5 & All Crowns\\
Takeshige et al. \cite{takeshige_high-resolution_2025} & temperate & natural  & 4.3k & 2.2--2.5 & Subset \\
SiDroForest \cite{vangeffenSiDroForestComprehensiveForest2022} & temperate & natural & 872 & 3 & Subset \\
Allen et al. \cite{allenLowcostTreeCrown2024} & temperate & natural & 2163 & 3 & All Crowns \\
OAM-TCD \cite{veitch-michaelis_oam-tcd_2024} & worldwide & mostly urban  &  280k & 10 & All Crowns \\
BCI50ha \cite{vasquezBarroColoradoIsland2023} & tropical & natural & 4.7k$^{\ast}$ & 4.5 & Subset \\
Detectree2 \cite{ballAccurateDelineationIndividual2023} & tropical & natural & 3.8k & 10 & Subset \\ 
\textbf{\textsc{SelvaMask} (ours)} & \textbf{tropical} & \textbf{natural} & \textbf{8.9k} & \textbf{1.3--3.5} & \textbf{All Crowns}\\
\bottomrule
\end{tabular}
\caption{\textbf{Comparison of publicly available tree crown segmentation datasets.} `Scope' indicates whether annotations aim to delineate all visible crowns in the imagery or a subset of trees (when stated by the dataset).$^{\ast}$BCI50ha reports crown maps across two dates; counts include repeated crowns for the same trees across time.}
\label{tab:related_datasets}
\vspace{-10pt}
\end{table*}

\section{Introduction}

Tropical forests sustain most of Earth's tree biodiversity \cite{gatti_number_2022} and carbon \cite{pan2011carbonsink}, with large canopy trees exerting disproportionate influence on carbon storage and ecosystem function \cite{lutz_global_2018}. Accurate crown segmentation from high-resolution remote sensing imagery enables biologists to study tree survival, growth, phenology or health \cite{araujo_integrating_2020}, facilitates tree species identification \cite{ball_towards_2024}, and serves as a proxy for aboveground biomass estimation \cite{jucker_tallo_2022}. However, individual tree crown segmentation in complex tropical canopies remains challenging due to dense foliage and overlapping crowns, high tree species diversity \cite{ballAccurateDelineationIndividual2023}, and high liana cover \cite{waite_view_2019}.

Tree crowns vary substantially in shape, color and texture (\eg, palms vs. broadleaf trees) and size distributions across biomes \cite{baudchon_selvabox_2026}, while aerial imagery across sites can differ in weather, lighting, resolution, and reconstruction quality. 
These distribution shifts and domain gaps in forest monitoring offer significant opportunities for application-driven machine learning innovation \cite{rolnick_position_2024}.
However, generalized individual tree crown segmentation remains underexplored, with most work focusing on specific regions \cite{ouaknineOpenForestDataCatalog2025}.

While tree crown segmentation has been primarily explored using airborne LiDAR \cite{xiang_forestformer3d_2025, holvoet_terrestrial_2025}, high-resolution RGB imagery offers a considerably cheaper and more accessible alternative for tropical researchers, motivating our RGB focus to facilitate broader adoption \cite{tagle_casapia_effective_2025}.
This focus enables leveraging recent VFMs based on RGB imagery, such as SAM \cite{kirillov_segment_2023}, demonstrating strong performance on instance segmentation tasks in low-data regimes. 

Off-the-shelf VFMs generate instance masks from prompts (\eg, bounding boxes), eliminating expensive manual annotations, a critical advantage for tree segmentation. This is particularly valuable given tropical forest dataset scarcity, though in-domain adaptation strategies may still be needed. 

Advancing tropical tree crown segmentation from RGB imagery requires diverse, high-quality datasets, which remain scarce (Tab.~\ref{tab:related_datasets}) despite the global importance of these forests. There is a clear need for new tropical datasets with dense annotations, high-resolution imagery, and diverse geographic coverage. Given the limited availability of annotations, adapting VFMs for generalized tree crown segmentation and comparing them to end-to-end approaches is essential. However, comprehensive benchmarking, including spatial cross-validation, out-of-distribution (OOD) evaluation, and domain-specific metrics, has yet to be conducted to assess current method limitations and guide future improvements.

Our contributions aim to overcome these challenges with:
\begin{itemize}
    \item \textbf{\SelvaMask}: The largest open tropical crown delineation dataset to date, spanning three Neotropical sites (Panama, Brazil, Ecuador) with higher-resolution imagery and denser manual annotations (Sec.~\ref{sec:expe_results}) than existing datasets (Tab.~\ref{tab:related_datasets}, Fig.~\ref{fig:qualitative_teaser}).
    \item \textbf{A comprehensive benchmark}: Evaluation of end-to-end and modular pipeline generalization capacities, including a new mRF1 metric, performance stratified by ecologically interpretable crown-size classes, and cross-site evaluation via spatial cross-validation.
    \item \textbf{State-of-the-art performance}: Fine-tuning VFMs with a detection-prompter achieves superior results as a modular approach, challenging the conventional end-to-end paradigm in both in-distribution and OOD settings, with potential for ecologically meaningful impact and broader adoption in tropical forests. 
    \item \textbf{Task ambiguity quantification}: Multi-annotator consistency analysis comparing model performance with human agreement (Sec.~\ref{sec:annot_agreement}), distinguishing genuine model failures from inherently ambiguous regions in dense tropical canopies.
\end{itemize}

\section{Related work}
\label{sec:related_work}

\paragraph{Datasets.} We summarize open-access datasets for individual tree crown segmentation in Table~\ref{tab:related_datasets}.

Corresponding datasets in natural tropical forests remain scarce, often limited to a single site or focused exclusively on larger trees, such as Detectree2 \cite{ballAccurateDelineationIndividual2023} and BCI50ha \cite{vasquezBarroColoradoIsland2023}. In contrast, \SelvaMask addresses these limitations by including multi-site rasters annotated under a complete-crown protocol (Fig.~\ref{fig:qualitative_teaser}, Sec.~\ref{app:annotation_guidelines}) capturing the dense, interlocking structure of the full canopy. 

By preserving small, crowded instances in diverse locations and carefully delineating boundaries, \SelvaMask offers twice the annotations of existing tropical datasets, establishing a substantially more challenging segmentation benchmark.

\paragraph{Evaluation protocols}
Tree detection benchmarks typically employ tile-wise classification metrics (precision, recall, F1) \cite{weinstein_benchmark_2021, santos_assessment_2019} or detection metrics such as intersection over union (IoU) \cite{haoAutomatedTreecrownHeight2021} and mean average precision (mAP) \cite{firozeTreeInstanceSegmentation2023, braga_tree_2020}.

However, tile-level evaluation introduces boundary artifacts and fails to accurately reflect performance on continuous forest rasters, the operational context for forest monitoring.
To address this in municipal tree inventories, \citet{veitch-michaelis_oam-tcd_2024} introduced a raster-level recall metric for keypoints.
Since keypoint recall ignores false positives and detection quality, \citet{baudchon_selvabox_2026} proposed a standardized IoU-based F1 score at the raster level (RF1), using Non-Maximum Suppression (NMS) to aggregate tile-level bounding box predictions.
However, this metric is limited by an individual IoU thresholds (focusing on 0.75, \ie RF1$_{75}$).
To address these limitations and align with mainstream COCO standards \cite{lin_microsoft_2014}, we adapt RF1 for instance segmentation and extend it to mRF1, averaging the F1 score over IoU thresholds 0.50--0.95 (Sec.~\ref{sec:eval_only}).

\paragraph{Architectures for tree crown segmentation.} 

Previous works have demonstrated the efficiency of U-Net \cite{ronneberger_u-net_2015} and Mask R-CNN \cite{he_mask_2017} architectures for binary \cite{chadwickAutomaticDelineationHeight2020,haoAutomatedTreecrownHeight2021, freudenbergIndividualTreeCrown2022, yangDetectingMappingTree2022, gongIndividualTreeAGB2023} or multiclass \cite{chadwickTransferabilityMaskRCNN2024, weishauptIndividualTreeCrown2025} tree crown segmentation. 
Despite strong detection capabilities, these models face challenges with precise boundary delineation, particularly in separating overlapping crowns within dense forest canopies.

Recently, transformer-based architectures such as Mask2Former \cite{cheng_masked-attention_2022} and Mask DINO \cite{li_mask_2023} have been adopted for their global attention mechanisms, enabling stronger contextual reasoning \cite{voulgarisBridgingClassicalModern, yangDiffKNetTLMaizePhenology2025}. 
Yet their application to dense tropical canopies with ambiguous boundaries remains unexplored. 

\paragraph{Generalized tree crown segmentation.}
While generalized tree detection has been addressed by models such as DeepForest \cite{zhouDeepForest2020} and SelvaBox \cite{baudchon_selvabox_2026}, equivalent generalization for instance segmentation remains underexplored. To our knowledge, Detectree2 \cite{ballAccurateDelineationIndividual2023}, a Mask R-CNN with ResNet-101 \cite{he_deep_2016} backbone pre-trained on diverse forest datasets including tropical sites, stands as the current state-of-the-art for generalized tropical tree crown segmentation.

\paragraph{Leveraging vision foundation models.} VFMs for instance segmentation, such as SAM-based approaches \cite{kirillov_segment_2023, raviSAM2Segment2024, carionSAM3Segment} show strong zero-shot performance across domains, though they fail to generalize effectively for tree crown segmentation \cite{tengBringingSAMNew2025}. 
Recent work combines domain-specific detectors like DeepForest \cite{zhouDeepForest2020} to prompt SAM-based approaches \cite{chenZeroShotTreeDetection2025, mengyuanTwoStageSegmentThenClassifyStrategy2025, queFMSAMIndividualTree2026, lungovaschettiTreePseCoScalingIndividual2025} avoiding the need for costly mask annotations but constraining generalization by relying on fixed, pretrained components.
Other methods enhance SAM backbones via learned prompts with RSPrompter \cite{chen_rsprompter_2024}, leveraging digital surface models with BalSAM \cite{tengBringingSAMNew2025} or use LiDAR point clouds as point-prompts \cite{zhuLeveragingSAM22025}.
To our knowledge, fine-tuning of both detection prompter from RGB imagery and VFM components for generalist tree crown segmentation remains unexplored.

\section{\SelvaMask dataset}
\label{sec:dataset}

\begin{figure}[t]
    \centering
    \includegraphics[width=\linewidth]{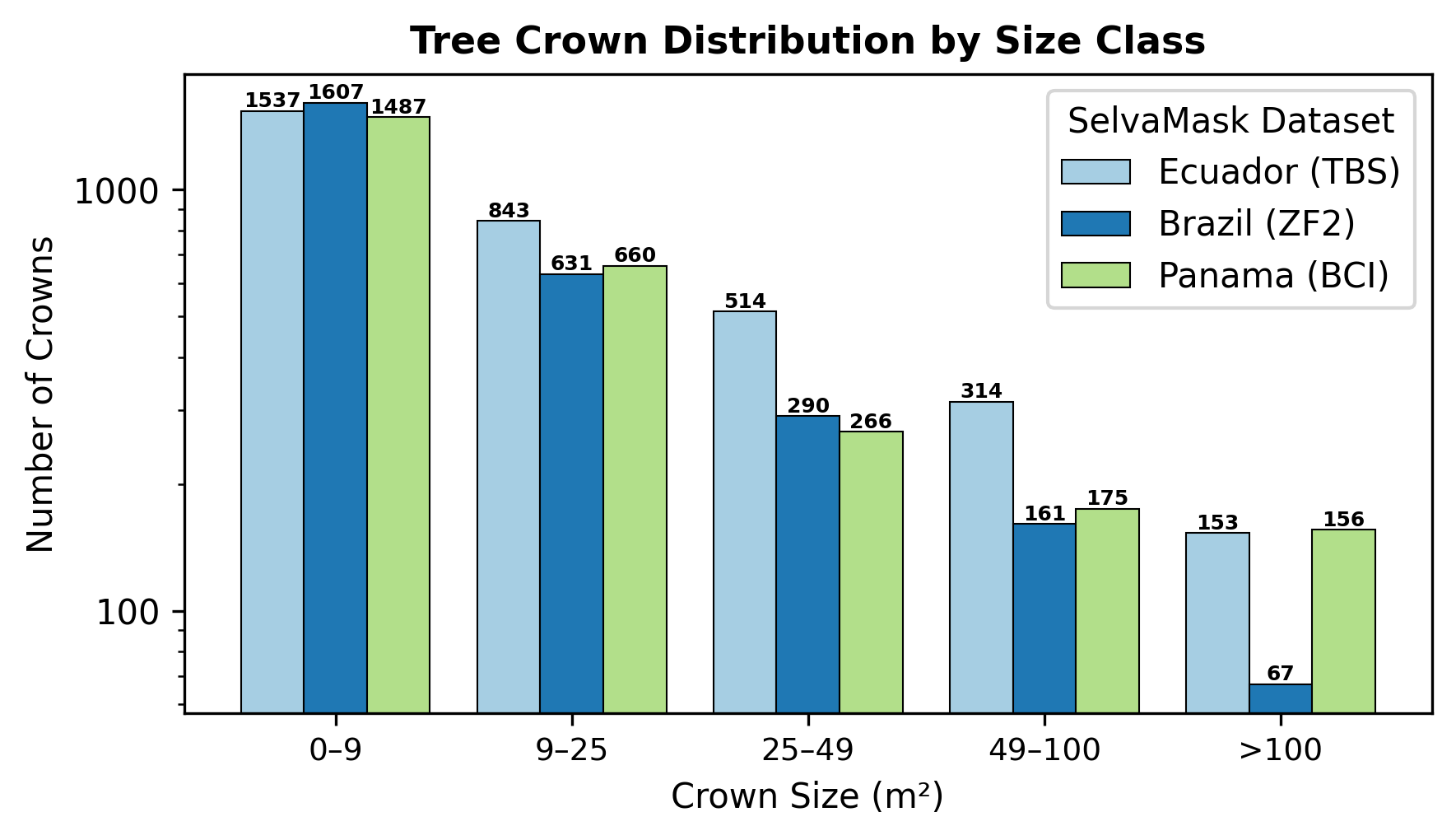}
    \caption{\textbf{Crown area distribution in \SelvaMask}. Vertical lines indicate the five ecological size classes (Tiny, Small, Medium, Large, Giant). Y-axis is in logarithmic scale.}
    \label{fig:size_dist}
    \vspace{-15pt}
\end{figure}

\SelvaMask is the largest tropical dataset for tree crown segmentation with 8\,861 manual annotations (Tab.~\ref{tab:related_datasets}), distributed in three Neotropical forest sites in Barro Colorado Island (Panama), ZF2 Reserve (Brazil) and Tiputini Biodiversity Station (Ecuador) as detailed in Table~\ref{tab:split_stats}.

\subsection{Image acquisition}
We acquired high-resolution orthomosaics across all sites using the DJI Mavic 3 Enterprise UAV, equipped with a 20MP 4/3 CMOS sensor. The imagery resolution spans between 1.3-3.5 cm per pixel, covering a total of 229.8 hectares. The imagery was captured under overcast conditions to minimize shadows and reduce contrast. 
Details on the recording process and orthomosaics are provided in Appendix~\ref{appendix:Image_Acquisition}. 

\subsection{Annotation protocol}
\label{sec:annotation_protocol}

\SelvaMask is designed for complete crown mapping: within each annotated tile, we delineate all visible tree crowns rather than only upper canopy or emergent individuals (see Fig.~\ref{fig:qualitative_teaser}). We define a crown instance as the canopy region visually attributable to a single tree in the orthomosaic, with boundaries following the visible outer contour. Non-tree objects and corrupted regions (\eg, poor photogrammetric reconstruction) were excluded.

An expert annotator manually traced crowns using ArcGIS Pro's Freehand Autocomplete tool to enforce strict topological consistency and prevent overlapping polygons. The detailed  procedure is provided in Appendix~\ref{app:annotation_guidelines}. Following initial tracing, the same annotator performed a secondary review to ensure consistent segmentation quality across sites, followed by an independent review by a different annotator.

\subsection{Dataset statistics} \label{sec:dataset_statistics}
We computed crown area in m$^2$ from polygons and orthomosaic ground sampling distance (1.3--3.5 cm/pixel). For ecologically interpretable analysis, we group crowns into five size classes: Tiny ($0$--$9$m$^2$), Small ($9$--$25$m$^2$), Medium ($25$--$49$m$^2$), Large ($49$--$100$m$^2$), and Giant ($>100$m$^2$). 

\SelvaMask captures the typical structure of tropical canopies, where many densely packed small crowns surround fewer emergent trees (Fig.~\ref{fig:qualitative_teaser}), a segmentation challenge not represented in prior tropical datasets. \SelvaMask is naturally dominated by small crowns: 52.3\% Tiny, 24.1\% Small, 12.1\% Medium, 7.3\% Large, and 4.2\% Giant (Fig.~\ref{fig:size_dist}). By contrast, current tropical datasets (Tab.~\ref{tab:related_datasets}) are biased toward large trees, with smaller trees ignored or overlooked during labelling (Appendix \ref{appendix:dataset_statistics}, Fig.~\ref{fig:size_dist_all_dataset}). A bias toward large trees is justifiable since they contribute disproportionately to carbon storage and forest functioning \cite{lutz_global_2018}. Nonetheless, the complete canopy coverage of \SelvaMask (Fig.~\ref{fig:qualitative_teaser}) allows us to stratify performance analysis by crown scale, from tiny to giant trees.

\subsection{Splits and tiling strategy}
To minimize spatial autocorrelation between data splits, we apply two-stage preprocessing: a zone-level partitioning of the forest area followed by a tile-level extraction.

\paragraph{Geospatial split.} 
We divide each orthomosaic into three disjoint zones (training, validation, test) with no overlapping trees (App.~\ref{appendix:split_zones}), enabling rigorous generalization assessment (Tab.~\ref{tab:split_stats}). 

\paragraph{Tile extraction.} 
Image tiles are extracted from each zone using a sliding window, with dimensions calibrated to capture the largest crowns without truncation. To prevent spatial autocorrelation, we mask pixels outside zone boundaries as transparent, ensuring strict spatial isolation between splits following the SelvaBox protocol \cite{baudchon_selvabox_2026}. Training tiles are extracted at $3555 \times 3555$ pixels
with 50\% overlap to support augmentation and maximize context, while validation and test use $1777 \times 1777$ pixels with 75\% overlap for standardized evaluation.

\begin{table}[t]
\centering
\resizebox{0.48\textwidth}{!}{%
\small
\setlength{\tabcolsep}{4pt}
\begin{tabular}{@{}l | cc | cc | cc@{}}
\toprule
& \multicolumn{2}{c|}{\textbf{Train}} & \multicolumn{2}{c|}{\textbf{Val}} & \multicolumn{2}{c}{\textbf{Test}} \\
\textbf{Site} & \textbf{Area (ha)} & \textbf{Crowns} & \textbf{Area (ha)} & \textbf{Crowns} & \textbf{Area (ha)} & \textbf{Crowns} \\
\midrule
Panama & 5.15 & 1499 & 1.53 & 570 & 2.93 & 783 \\
Brazil & 3.31 & 1590 & 0.92 & 568 & 1.16 & 716 \\
Ecuador & 7.20 & 1836 & 1.89 & 629 & 2.91 & 896 \\
\midrule
\textbf{Total} & \textbf{15.7} & \textbf{4925} & \textbf{4.3} & \textbf{1767} & \textbf{7.0} & \textbf{2395} \\
\bottomrule
\end{tabular}
}
\caption{\textbf{Dataset distribution.} Annotated area (in hectares) and individual crown counts across geographic sites and splits.}
\label{tab:split_stats}
\vspace{-20pt}
\end{table}

\subsection{Inter-annotator agreement}
\label{sec:annot_agreement}
To quantify the ambiguity of tropical crown delineation, we conducted an inter-annotator agreement study on \SelvaMask. 
We select subsets from the test zones across all three sites, totaling approximately 500 crowns. Two additional annotators independently traced all visible crowns in these regions following the protocol in Section~\ref{sec:annotation_protocol}. 
We computed pairwise agreement scores with each annotator serving alternately as reference, establishing a human-level performance baseline.

As shown in Table~\ref{tab:detailed_per_gt_comparison} and Section~\ref{sec:expe_results}, inter-annotator agreement decreases with crown size, revealing an expert-level performance ceiling for distinguishing small, overlapping crowns, an open challenge in this domain. The results also expose annotator-specific size biases, with some systematically favoring smaller crowns and others larger crowns.

\section{Methods and evaluation protocol}
\label{sec:method_and_eval}

We address the instance segmentation task: given an RGB orthomosaic tile, models must detect all tree crowns and produce binary segmentation masks for each instance. 

\subsection{End-to-end competing methods}
We evaluate standard Mask R-CNN \cite{he_mask_2017} and Mask2Former \cite{cheng_masked-attention_2022} architectures. We use a ResNet-50 \cite{he_deep_2016} backbone for both architectures, and Swin-L \cite{liu_swin_2021} for Mask2Former. These serve as our primary \textit{end-to-end baselines}, pretrained on COCO and fine-tuned on \SelvaMask, to establish the upper bound of traditional supervision.

We include RSPrompter \cite{chen_rsprompter_2024}, a prompt-learning framework for remote sensing. We train it end-to-end on \SelvaMask to assess state-of-the-art domain-specific methods \cite{tengBringingSAMNew2025}.
RSPrompter's $1024 \times 1024$ input constraint necessitates resizing larger tiles while maintaining the same data augmentation pipeline for benchmark consistency. This downsampling reduces spatial resolution and likely degrades segmentation performance, representing a limitation of RSPrompter on our benchmark.

Finally, we evaluate two pretrained checkpoints of Detectree2 \cite{ballAccurateDelineationIndividual2023}: 
the \textit{resize} and \textit{flexi} models, trained on standard and urban-augmented Detectree2 data, respectively. 
We select Detectree2 as the primary competing method as it represents the leading approach for generalized tree crown segmentation.
We evaluate it zero-shot to benchmark the OOD generalization capability of existing generalist models on our benchmark.

\subsection{Detection-segmentation modular pipelines}
\label{sec:modular_pipeline}

\paragraph{Modular approach.}
We propose to approach the tree crown segmentation task by prompting SAM-based VFMs with a detection model. We will denote this pipeline as `\textbf{detection-prompter} $\rightarrow$ \textbf{SAM}' in this work, where \textbf{SAM} can be version 2 or 3.
We evaluate three \textbf{detection-prompter} variants: DeepForest \cite{zhouDeepForest2020}, a tree detection model; DINO \cite{caron_emerging_2021}, a DINO-Swin-L backbone pre-trained on COCO; and SelvaBox \cite{baudchon_selvabox_2026}, a DINO-Swin-L detection model trained on tropical forest data. 
Both DeepForest and SelvaBox are generalist tree detection models, considered as domain-specific detection-prompter in our pipeline, making them well-suited to our evaluation framework.

For all `detection-prompter $\rightarrow$ SAM' pipelines, each predicted box is passed as a box prompt to SAM ensuring one mask per box. The resulting mask is treated as one crown instance, with combined detection-segmentation confidence as its score. Each pipeline outputs up to 300 predicted crown instances per tile, evaluated using COCO-style instance segmentation metrics with IoU-based matching between predictions and ground truth.

\paragraph{Pipeline fine-tuning.}
Since promptable VFMs are bounded by the quality of their prompts \cite{tengBringingSAMNew2025}, we first fine-tune detection models using \SelvaMask's segmentation masks converted to bounding boxes. This yields the strongest possible proposal mechanism. We then select this best-performing detection-prompter for all SAM experiments to isolate the impact of adaptation.

We perform end-to-end fine-tuning of SAM2 and SAM3, updating the image encoder, prompt encoder, and mask decoder jointly.
To simulate realistic inference, we use predicted boxes from the detection-prompter rather than ground-truth annotations.
Our loss function combines Dice and IoU prediction terms:
$\mathcal{L} = \mathcal{L}_{\text{dice}} + \mathcal{L}_{\text{iou}}$.
We omit the standard Focal loss ($\lambda_{\text{focal}}=0$) as validation experiments showed it degraded performance on our dataset (Sec.~\ref{sec:results_selvamask}).

Full training details are provided in Appendix~\ref{appendix:Hyperparameters}.

\subsection{Evaluation}
\label{sec:eval_only}

\paragraph{COCO-style tile-level evaluation.}
Predictions are compared to ground-truth crown masks using IoU-based one-to-one matching: each prediction can match at most one ground-truth crown and vice versa. Unmatched predictions are counted as false positives, while unmatched ground-truth crowns are false negatives. 
We report COCO-style instance segmentation metrics, namely mean average precision (mAP), mean average recall (mAR) (averaged over IoU $0.50:0.95$), mAP$_{50}$, and mAP$_{75}$, which robustly evaluate segmentation across the many crown instances per tile.
Metrics are aggregated over all test tiles and stratified by crown size (Tiny to Giant; Sec.~\ref{sec:dataset_statistics}).

\begin{table*}[t]
\centering
\scriptsize
\setlength{\tabcolsep}{2pt}
\renewcommand{\arraystretch}{1.2}
\setlength{\aboverulesep}{0pt}
\setlength{\belowrulesep}{0pt}

\begin{tabular}{@{}cl:c:ccc:ccc:ccc@{}}
\toprule
& \textbf{Method} & \textbf{OOD} & \textbf{mAP} & \textbf{AP$_{50}$} & \textbf{AP$_{75}$} & \textbf{mAR} & \textbf{AR$_{50}$} & \textbf{AR$_{75}$} & \textbf{mRF1} & \textbf{RF1$_{50}$} & \textbf{RF1$_{75}$} \\
\midrule

& Detectree2 (flexi)\ice & \cmark & 7.8 & 18.9 & 5.2 & 12.3 & 26.5 & 10.2 & 15.7 & 33.8 & 12.7 \\
& Detectree2 (resize)\ice & \cmark & 5.0 & 11.7 & 3.5 & 9.8 & 21.7 & 7.7 & 5.6 & 10.2 & 5.6 \\
& Mask R-CNN (ResNet-50)\fire & \xmark & 11.1\tiny($\pm1.7$) & 28.7\tiny($\pm3.5$) & 6.6\tiny($\pm1.5$) & 21.6\tiny($\pm1.9$) & 48.9\tiny($\pm3.8$) & 16.3\tiny($\pm1.7$) & 17.8\tiny($\pm1.9$) & 38.0\tiny($\pm3.4$) & 14.8\tiny($\pm2.1$) \\
& Mask2Former (ResNet-50)\fire & \xmark & 10.5\tiny($\pm0.3$) & 24.8\tiny($\pm0.5$) & 7.5\tiny($\pm0.5$) & 16.0\tiny($\pm0.1$) & 33.3\tiny($\pm0.0$) & 13.9\tiny($\pm0.3$) & 23.1\tiny($\pm0.6$) & 49.0\tiny($\pm0.7$) & 19.8\tiny($\pm1.0$) \\
\rowcolor{gray!15}
\cellcolor{white} & Mask2Former (Swin-L)\fire & \xmark & 19.7\tiny($\pm0.9$) & 42.4\tiny($\pm1.1$) & 16.2\tiny($\pm1.0$) & 26.5\tiny($\pm0.9$) & 52.8\tiny($\pm0.9$) & 23.9\tiny($\pm1.2$) & 30.8\tiny($\pm1.3$) & 57.3\tiny($\pm2.5$) & 29.7\tiny($\pm1.2$) \\
& RSPrompter (SAM-base)\fire & \xmark & 6.6\tiny($\pm0.8$) & 15.5\tiny($\pm1.5$) & 4.9\tiny($\pm0.7$) & 10.2\tiny($\pm0.7$) & 21.8\tiny($\pm1.3$) & 8.6\tiny($\pm0.6$) & 13.4\tiny($\pm0.9$) & 28.6\tiny($\pm1.7$) & 11.2\tiny($\pm0.7$) \\
\multirow{-6}{*}{\rotatebox[origin=c]{90}{\textit{End-to-end}}} 
& RSPrompter (SAM-huge)\fire & \xmark & 16.3\tiny($\pm0.3$) & 35.1\tiny($\pm0.5$) & 13.9\tiny($\pm0.3$) & 21.3\tiny($\pm0.4$) & 41.6\tiny($\pm0.6$) & 20.0\tiny($\pm0.5$) & 26.5\tiny($\pm0.0$) & 50.0\tiny($\pm0.5$) & 26.1\tiny($\pm0.4$) \\
\hdashline

& SelvaBox\ice$\rightarrow$SAM2\ice & \cmark & 16.5 & 33.2 & 14.6 & 30.6 & 60.0 & 28.2 & 29.6 & 56.2 & 28.3 \\
& SelvaBox\fire$\rightarrow$SAM2\ice & \xmark & 20.9 & 44.3 & 17.3 & 35.0 & 71.7 & 30.5 & 32.1 & 61.7 & 30.9 \\
\rowcolor{gray!15}
\cellcolor{white}\multirow{-3}{*}{\rotatebox[origin=c]{90}{\textit{SAM2}}} 
& SelvaBox\fire$\rightarrow$SAM2\fire & \xmark & \underline{23.6\tiny($\pm0.2$)} & \textbf{48.3\tiny($\pm0.6$)} & \underline{21.1\tiny($\pm0.2$)} & \underline{37.1\tiny($\pm0.3$)} & 72.0\tiny($\pm0.2$) & \underline{33.8\tiny($\pm0.3$)} & \underline{35.3\tiny($\pm0.3$)} & \underline{63.5\tiny($\pm0.6$)} & \underline{35.3\tiny($\pm0.5$)} \\
\hdashline

& DeepForest\ice$\rightarrow$SAM3\ice & \cmark & 5.3 & 12.6 & 3.7 & 7.3 & 16.3 & 5.9 & 12.6 & 28.5 & 9.7 \\
& DINO\fire$\rightarrow$SAM3\ice & \xmark & 20.5 & 42.9 & 17.1 & 35.1 & 70.8 & 30.7 & 32.4 & 61.8 & 30.1 \\
& SelvaBox\ice$\rightarrow$SAM3\ice & \cmark & 17.2 & 34.4 & 15.4 & 31.7 & 61.1 & 29.5 & 30.2 & 57.3 & 29.6 \\
& SelvaBox\fire$\rightarrow$SAM3\ice & \xmark & 22.0 & 45.9 & 18.5 & 36.5 & \underline{73.1} & 32.1 & 32.8 & 62.0 & 31.7 \\
\rowcolor{gray!15}
\cellcolor{white}\multirow{-5}{*}{\rotatebox[origin=c]{90}{\textit{SAM3}}} 
& SelvaBox\fire$\rightarrow$SAM3\fire & \xmark & \textbf{24.4\tiny($\pm0.1$)} & \underline{46.9\tiny($\pm0.3$)} & \textbf{22.5\tiny($\pm0.1$)} & \textbf{38.7\tiny($\pm0.0$)} & \textbf{73.1\tiny($\pm0.1$)} & \textbf{36.4\tiny($\pm0.1$)} & \textbf{36.3\tiny($\pm0.1$)} & \textbf{63.9\tiny($\pm0.4$)} & \textbf{36.9\tiny($\pm0.4$)} \\
\bottomrule
\end{tabular}
\caption{\textbf{Quantitative comparison on \SelvaMask.} We compare end-to-end and modular SAM-based instance segmentation methods (see Sec.~\ref{sec:method_and_eval}). Best and second-best scores overall are marked in \textbf{bold} and \underline{underlined}. {\setlength{\fboxsep}{1pt}\colorbox{gray!15}{Gray rows}} denote the best performance within each section. `OOD' indicates whether the model is evaluated out-of-distribution (zero-shot) (\cmark) or was trained on the target domain (\xmark). We note `\fire' models that are fine-tuned on \SelvaMask and `\ice' models whose weights are kept frozen.}
\label{tab:main_merged_results}
\vspace{-10pt}
\end{table*}

\paragraph{Raster-level evaluation.}

To evaluate instance segmentation on full orthomosaics, we propose the mean Raster-level F1 (mRF1), defined as
\setlength{\abovedisplayskip}{0pt}
\setlength{\belowdisplayskip}{0pt}
\setlength{\abovedisplayshortskip}{0pt}
\setlength{\belowdisplayshortskip}{0pt}
$$\text{mRF1} = \frac{1}{|\mathcal{T}|}\sum_{\tau \in \mathcal{T}} \text{RF1}_{\tau},$$
where $\mathcal{T} = \{0.50, 0.55, \dots, 0.95\}$ represents the set of IoU thresholds. This aggregation captures both global counting accuracy and local segmentation fidelity.
We further extend the original RF1 methodology \cite{baudchon_selvabox_2026} by stratifying scores into the same five size classes used for our COCO-style metrics (Tiny to Giant; Sec.~\ref{sec:dataset_statistics}).
This enables a novel analysis of model performance across different crown scales by filtering predictions and ground truth based on area, mirroring the standard COCO evaluation protocol \cite{lin_microsoft_2014}.
Calculating the mRF1 metric requires optimizing NMS and confidence thresholds on a validation set prior to benchmarking on the test set; we detail this optimization protocol in Appendix~\ref{appendix:nms_params}.

\paragraph{OOD evaluations.}
We perform a spatial cross-validation on \SelvaMask to assess generalization capacities that sites could provide to an unseen region while motivating their complementarity. We train Mask2Former (Swin-L) on two sites and evaluate it zero-shot on the third considered as OOD (Sec.~\ref{sec:results_ood}, Tab.~\ref{tab:cross_site_results}).
To assess the generalization capacities of all methods, we evaluated frozen and fine-tuned  competing and proposed models with \SelvaMask on external datasets, either in-distribution or OOD, from different biomes worldwide (Sec.~\ref{sec:results_ood}, Tab.~\ref{tab:ood_results}).

\section{Experiments and results}
\label{sec:expe_results}

\subsection{Experimental setup}
For all trained or fine-tuned methods, we include data augmentations such as random resized crops, flips, and rotations, including the full list in Appendix~\ref{appendix:data_augmentation}.
Training and inference hyperparameters used for each model are detailed in Appendix~\ref{appendix:Hyperparameters}.
Fine-tuning, training and evaluation experiments used NVIDIA H100 GPUs. Depending on the model, training sessions ran from 6 hours to 3 days. 

\paragraph{External datasets.}
Beyond our \SelvaMask benchmark, we conduct an ablation study on five additional datasets (Tab.~\ref{tab:ood_results}): two tropical (Detectree2, BCI50ha), two temperate (BAMForest, QuebecTrees), and one mixed-biome urban (OAM-TCD). Since SelvaBox was trained on QuebecTrees and OAM-TCD, our SelvaBox-based methods are in-distribution on these two benchmarks, unlike DeepForest and Detectree2 which were not trained on them. All methods remain OOD on BCI50ha, BAMForest, and the Detectree2 dataset. 

\begin{table*}[t]
\centering
\scriptsize
\setlength{\tabcolsep}{2pt}
\renewcommand{\arraystretch}{1.2}
\setlength{\aboverulesep}{0pt}
\setlength{\belowrulesep}{0pt}
\resizebox{0.90\textwidth}{!}{%
\begin{tabular}{@{}c: l:l:c:c:ccccc:c:ccccc@{}}
\toprule
& & & & \multicolumn{6}{c:}{\textbf{mAP}} & \multicolumn{6}{c}{\textbf{mRF1}} \\
\cmidrule(lr){5-10} \cmidrule(lr){11-16}
& \textbf{Dataset} & \textbf{Method} & \textbf{OOD} & \textbf{All} & \textbf{Tiny} & \textbf{Small} & \textbf{Medium} & \textbf{Large} & \textbf{Giant} & \textbf{All} & \textbf{Tiny} & \textbf{Small} & \textbf{Medium} & \textbf{Large} & \textbf{Giant} \\
\midrule
\multirow{10}{*}{\rotatebox[origin=c]{90}{\textit{Tropical}}}
& \multirow{5}{*}{\textbf{Detectree2}} 
& Detectree2 (flexi) & \xmark &17.3 & 0.7 & 2.9 & 6.0 & \underline{16.6} & \underline{28.0} & 21.9 & 0.0 & 0.9 & 4.6 & 15.9 & 31.6 \\
& & Detectree2 (resize) & \xmark & 9.8 & 0.0 & 1.6 & 4.1 & 12.2 & 16.0 & 14.9 & 0.0 & 0.0 & 0.3 & 8.1 & 23.2 \\
& & DeepForest\ice$\rightarrow$SAM3\ice & \cmark & 11.4 & 0.8 & 5.6 & 7.5 & 13.0 & 17.0 & 18.1& \textbf{3.5} & 9.4 & 11.3 & 15.6 & 21.4 \\
& & SelvaBox\ice $\rightarrow$ SAM3\ice & \cmark & \underline{17.7} & \underline{1.3} & \underline{5.7} & \underline{9.9} & 16.3 & 27.7 & \underline{23.6} & 1.3 & \underline{9.6} & \underline{15.2} & \underline{22.0} & \underline{37.8} \\
& & SelvaBox\fire $\rightarrow$ SAM3\fire & \cmark & \textbf{21.6($\pm0.3$)} & \textbf{1.9($\pm0.1$)} & \textbf{9.7($\pm0.4$)} & \textbf{12.0\tiny($\pm0.5$)} & \textbf{20.2\tiny($\pm0.1$)} & \textbf{33.0\tiny($\pm0.5$)} &\textbf{26.0($\pm0.4$)} & \underline{2.4($\pm0.2$)} & \textbf{11.2}($\pm0.4$) & \textbf{17.6\tiny($\pm0.2$)} & \textbf{24.9\tiny($\pm0.2$)} & \textbf{39.8\tiny($\pm0.2$)} \\
\cdashline{2-16} 
& \multirow{5}{*}{\textbf{BCI50ha}} 
& Detectree2 (flexi) & \cmark & 29.3 & 0.3 & 2.9 & \underline{19.9} & 36.2 & 40.8 & \textbf{36.6} & 0.0 & \textbf{6.8} & \textbf{23.7} & 35.5 & 41.5 \\
& & Detectree2 (resize) & \cmark & 26.5 & 0.0 & 1.6 & 14.3 & 29.8 & 35.9 & \underline{35.4} & 0.0 & 0.0 & 12.5 & 28.8 & 41.7 \\
& & DeepForest\ice$\rightarrow$SAM3\ice & \cmark & 20.3 & \textbf{0.4} & 1.7 & 14.7 & 26.7 & 30.0 & 27.0& \textbf{2.5} & 3.4 & 18.2 & 31.0 & 36.5 \\
& & SelvaBox\ice $\rightarrow$ SAM3\ice & \cmark & \underline{34.0} & \textbf{0.4} & \underline{3.2} & 18.8 & \underline{37.1} & \underline{51.3} & 25.2 & 0.3 & 3.7 & 21.2 & \underline{44.7} & \underline{60.5} \\
& & SelvaBox\fire $\rightarrow$ SAM3\fire & \cmark &\textbf{37.5($\pm0.7$)} & 0.2($\pm0.1$) & \textbf{4.8}($\pm0.3$) & \textbf{24.0\tiny($\pm0.6$)} & \textbf{41.3\tiny($\pm0.6$)} & \textbf{52.3\tiny($\pm0.7$)} & 31.7($\pm0.4$) & \underline{0.4($\pm0.0$)} & \underline{5.1($\pm0.1$)} & \underline{23.1\tiny($\pm0.2$)} & \textbf{45.6\tiny($\pm0.9$)} & \textbf{61.1\tiny($\pm0.5$)} \\
\cdashline{2-16}
\midrule
\multirow{10}{*}{\rotatebox[origin=c]{90}{\textit{Temperate}}}
& \multirow{5}{*}{\textbf{BAMForest}} 
& Detectree2 (flexi) & \cmark & 17.9 & 7.3 & 22.1 & 27.4 & 25.9 & 23.5 & -- & -- & -- & -- & -- & -- \\
& & Detectree2 (resize) & \cmark & 16.0 & 6.1 & 19.5 & 25.6 & 24.2 & 24.6 & -- & -- & -- & -- & -- & -- \\
& & DeepForest\ice$\rightarrow$SAM3\ice & \cmark & 12.0 & 6.5 & 17.3 & 18.3 & 9.6 & 4.0 & -- & -- & -- & -- & -- & -- \\
& & SelvaBox\ice $\rightarrow$ SAM3\ice & \cmark & \textbf{23.0} & \underline{9.0} & \underline{29.6} & \textbf{35.9} & \textbf{33.0} & \textbf{29.6} & -- & -- & -- & -- & -- & -- \\
& & SelvaBox\fire $\rightarrow$ SAM3\fire & \cmark & \underline{22.8}($\pm0.4$)& \textbf{10.6}\tiny($\pm0.2$) & \textbf{30.6}\tiny($\pm0.4$) & \underline{34.5}\tiny($\pm0.2$) & \underline{29.5}\tiny($\pm0.7$) & \underline{25.5}\tiny($\pm1.4$) & -- & -- & -- & -- & -- & -- \\
\cdashline{2-16}
& \multirow{5}{*}{\textbf{QuebecTrees}} 
& Detectree2 (flexi) & \cmark & 10.8 & 0.5 & 24.1 & 28.2 & 40.6 & 52.6 & 20.5 & 12.1 & 26.2 & 25.5 & 25.5 & 2.4 \\
& & Detectree2 (resize) & \cmark & 3.9 & 1.6 & 13.0 & 18.3 & 32.3 & 45.3 & 10.4 & 2.8 & 16.4 & 17.0 & 12.3 & 3.7 \\
& & DeepForest\ice$\rightarrow$SAM3\ice & \cmark & 8.3 & 5.0 & 17.3 & 16.7 & 25.9 & 42.3 & 17.1 & 11.4 & 18.7 & 11.6 & 18.4 & 0.0 \\
& & SelvaBox\ice $\rightarrow$ SAM3\ice & \xmark & \underline{25.8} & \underline{20.7} & \underline{36.8} & \underline{42.1} & \textbf{53.3} & \textbf{62.8} & \underline{38.1} & \underline{32.5} & \underline{38.8} & \underline{36.4} & \underline{41.1} & \underline{12.6} \\
& & SelvaBox\fire $\rightarrow$ SAM3\fire & \xmark & \textbf{27.6}($\pm0.3$) & \textbf{21.9\tiny($\pm0.3$)} & \textbf{40.8\tiny($\pm0.4$)} & \textbf{46.3\tiny($\pm0.5$)} &  \underline{51.6}\tiny($\pm1.5$) &  \underline{54.5}\tiny($\pm4.8$) & \textbf{39.8\tiny($\pm0.2$)} & \textbf{33.5\tiny($\pm0.1$)} & \textbf{45.0\tiny($\pm0.4$)} & \textbf{45.7\tiny($\pm0.4$)} & \textbf{49.9\tiny($\pm0.7$)} & \textbf{40.3\tiny($\pm0.0$)} \\
\midrule 
\multirow{5}{*}{\rotatebox[origin=c]{90}{\textit{Urban}}}   
& \multirow{5}{*}{\textbf{OAM-TCD}} 
& Detectree2 (flexi) & \cmark & \underline{12.3} & 0.7 & 7.4 & \underline{27.2} & \textbf{36.5} & \textbf{37.8} & -- & -- & -- & -- & -- & -- \\
& & Detectree2 (resize) & \cmark & 2.3 & 0.2 & 1.3 & 5.2 & 14.0 & 20.2 & -- & -- & -- & -- & -- & -- \\
& & DeepForest\ice$\rightarrow$SAM3\ice & \cmark & 6.5 & 0.2 & 6.8 & 16.3 & 16.5 & 12.4 & -- & -- & -- & -- & -- & -- \\
& & SelvaBox\ice $\rightarrow$ SAM3\ice & \xmark & 12.2 & \underline{6.9} & \underline{14.4} & 19.5 & 18.5 & 15.3 & -- & -- & -- & -- & -- & -- \\
& & SelvaBox\fire $\rightarrow$ SAM3\fire & \xmark & \textbf{18.5}($\pm1.0$) & \textbf{8.9($\pm0.9$)} & \textbf{21.9($\pm1.3$)} & \textbf{29.4($\pm1.7$)} & \underline{29.8($\pm1.4$)} & \underline{24.9($\pm1.6$)} & -- & -- & -- & -- & -- & -- \\
\bottomrule
\end{tabular}%
}
\caption{\textbf{Performance on external datasets.} Metrics are stratified by crown size class. `OOD' indicates whether the model is evaluated out-of-distribution (zero-shot) (\cmark) or was trained on the target domain (\xmark). `--' denotes not available mRF1 for temperate datasets lacking full raster annotations. We note `\fire' models that are fine-tuned on \SelvaMask and `\ice' models whose weights are kept frozen.}
\label{tab:ood_results}
\vspace{-10pt}
\end{table*}

\subsection{Results on \SelvaMask}
\label{sec:results_selvamask}

\paragraph{Modular pipelines lead the \SelvaMask benchmark.} We summarize our experiments on \SelvaMask in Table~\ref{tab:main_merged_results} distinguishing end-to-end methods and modular approaches, either with SAM2 or SAM3.
The proposed modular pipeline SelvaBox\fire $\rightarrow$ SAM3\fire, including both modules fine-tuned on \SelvaMask reaches the best performance overall outperforming the best end-to-end model (Mask2Former--Swin-L), also fine-tuned on \SelvaMask, by a large margin.
Off-the-shelf tropical models generalize poorly to dense, complete-crown mapping: Detectree2 and DeepForest$\rightarrow$SAM3 achieve single-digit mAP (7.9 and 6.2, respectively).
Overall, accurate tree crown segmentation in dense tropical canopies benefits from decoupling detection and segmentation, where our modular approach enables substantial gains over end-to-end models.

\paragraph{Better segmentation for larger crowns.}
Segmentation performance improves with crown dimensions (Appendix~\ref{appendix:additional_results}, Fig~\ref{fig:map_by_size}), reflecting clearer boundaries for large trees and greater ambiguity for small, overlapping canopies. Our fine-tuned modular pipeline excels across all size classes, particularly on `Giant' crowns (65.4 mAP vs. 40.4 mAP for Detectree2), while also showing substantial gains on `Tiny' and `Small' crowns, enabling more complete canopy census.

\paragraph{Frozen modular pipelines compete with fine-tuned end-to-end models.} Leveraging SelvaBox as a detection-prompter of SAM2 or SAM3 shows great generalization capacities without fine-tuning, as they outperform both versions of Detectree2 in an OOD setting.
The best frozen modular pipeline, SelvaBox\ice $\rightarrow$ SAM3\ice, achieves comparable performance to the best fine-tuned Mask2Former (Swin-L), demonstrating the effectiveness of modular pipelines with generalist in-domain detector-prompters.

\paragraph{Transformers lead among end-to-end methods}
Mask2Former with a Swin-L backbone achieves the highest performance among end-to-end models (19.7 mAP), significantly outperforming the CNN-based Mask R-CNN (11.1 mAP). This establishes Mask2Former as a promising end-to-end architecture for forest monitoring applications. 

\paragraph{Prompt quality drives performance.}
A dominant factor for promptable segmentation models is the quality of box prompts.
When using frozen SAM, switching from DeepForest to a more performant detection-prompter yields a near three-fold improvement, increasing mAP from 6.2 to 17.3 (Tab.~\ref{tab:main_merged_results}). 
This highlights that tropical crown delineation is primarily limited by localization under heavy canopy crowding, and that robust proposal mechanisms are essential for leveraging VFMs in this domain.

\paragraph{SAM fine-tuning is key.}
Fine-tuning consistently improves segmentation performance on SelvaMask, even when starting from strong pretrained backbones.
In particular, adapting SAM to tropical crown boundaries yields clear gains over its frozen counterpart under the same prompting protocol (Tab.~\ref{tab:main_merged_results}).
We conducted an ablation study on the specific components of SAM2 to fine-tune (Appendix~\ref{appendix:SAM_components}, Tab. \ref{tab:ablation_sam_components}) and found that full end-to-end adaptation was necessary to obtain optimal performances. These improvements are reflected in both COCO-style tile metrics (mAP/mAR) and raster-level fidelity (mRF1), indicating that fine-tuning helps to better capture fine-scale crown boundaries and reduce common failure modes in dense tropical canopies.

\paragraph{Models reach human performance on a difficult task.}
We further quantify task ambiguity by comparing model performance against inter-annotator agreement (Tab.~\ref{tab:detailed_per_gt_comparison}).
We first observe that pairwise expert agreement scores decline as crown size decreases (Sec.~\ref{sec:annot_agreement}), ranging from relatively high agreement (mRF1 58.1--72.8) for giant trees, down to marginal agreement (12.7--20.1) for tiny ones.
The lower agreement scores of small trees underscore the inherent difficulty of delineating crowns in dense tropical canopies. Crucially, our best model achieves scores comparable to, and often exceeding, pairwise human agreement. This implies that our pipeline has effectively reached the `human-consensus ceiling' for this dataset; remaining `errors' largely reflect valid interpretational differences rather than model failure. Consequently, surpassing this performance bound would likely require a refinement in annotation methodology, such as multi-expert consensus voting for low-agreement score trees, to resolve ground-truth ambiguity.

\begin{table*}[t]
\centering
\scriptsize
\setlength{\tabcolsep}{1.5pt}
\renewcommand{\arraystretch}{1.2}
\setlength{\aboverulesep}{0pt}
\setlength{\belowrulesep}{0pt}
\begin{tabular}{@{}l : cccc : cccc : cccc : c@{}}
\toprule
\multirow{2}{*}{\textbf{Method}} & \multicolumn{4}{c:}{\textbf{Brazil}} & \multicolumn{4}{c:}{\textbf{Ecuador}} & \multicolumn{4}{c:}{\textbf{Panama}} & \multicolumn{1}{c}{\textbf{All}} \\
\cmidrule(lr){2-5} \cmidrule(lr){6-9} \cmidrule(lr){10-13} \cmidrule(lr){14-14}
 & \textbf{mAP} & \textbf{mAR} & \textbf{mRF1} & \textbf{OOD} & \textbf{mAP} & \textbf{mAR} & \textbf{mRF1} & \textbf{OOD} & \textbf{mAP} & \textbf{mAR} & \textbf{mRF1} & \textbf{OOD} 

 & \textbf{mRF1} \\
\midrule
Mask2Former (Swin-L)\fire & 14.3\tiny($\pm0.4$) & 19.5\tiny($\pm0.5$) & 24.1\tiny($\pm0.5$) & \cmark & 21.6\tiny($\pm0.2$) & 30.2\tiny($\pm0.3$) & 31.9\tiny($\pm0.2$) & \xmark & \underline{18.5}\tiny($\pm0.1$) & \underline{25.1}\tiny($\pm0.1$) & \underline{29.1}\tiny($\pm0.2$) & \xmark

& 28.7\tiny($\pm0.2$) \\ 
Mask2Former (Swin-L)\fire & \underline{15.3}\tiny($\pm0.2$) & 20.7\tiny($\pm0.2$) & 26.6\tiny($\pm0.4$) & \xmark & 16.8\tiny($\pm0.2$) & 26.0\tiny($\pm0.3$) & 25.9\tiny($\pm0.3$) & \cmark & 17.6\tiny($\pm0.1$) & 24.3\tiny($\pm0.1$) & 28.0\tiny($\pm0.5$) & \xmark 

& 26.8\tiny($\pm0.2$) \\ 
Mask2Former (Swin-L)\fire & \textbf{16.7}\tiny($\pm0.3$) & \textbf{22.3}\tiny($\pm0.2$) & \textbf{29.0}\tiny($\pm0.6$) & \xmark & \underline{22.3}\tiny($\pm0.6$) & \textbf{31.0}\tiny($\pm0.6$) & \underline{32.6}\tiny($\pm1.0$) & \xmark & 17.4\tiny($\pm0.3$) & 24.7\tiny($\pm0.3$) & 27.9\tiny($\pm0.8$) & \cmark & 

\underline{30.0}\tiny($\pm0.8$) \\ 
\hdashline
Mask2Former (Swin-L)\fire & \textbf{16.7}\tiny($\pm0.7$) & \underline{21.9}\tiny($\pm0.8$) & \underline{28.6}\tiny($\pm1.6$) & \xmark & \textbf{22.8}\tiny($\pm1.1$) & \textbf{31.0}\tiny($\pm1.1$) & \textbf{33.5}\tiny($\pm1.3$) & \xmark & \textbf{18.6}\tiny($\pm0.9$) & \textbf{25.4}\tiny($\pm0.9$) & \textbf{29.7}\tiny($\pm1.2$) & \xmark 
& \textbf{30.8}\tiny($\pm1.3$) \\

\bottomrule
\end{tabular}
\caption{\textbf{Spatial cross-validation results.} Performance of Mask2Former (SwinL) on each \SelvaMask location, fine-tuned with an out-of-distribution location (`OOD'), compared to all locations (last row). We mark the best and second-best scores in \textbf{bold} and \underline{underline}.}
\label{tab:cross_site_results}
\vspace{-10pt}
\end{table*}
\begin{table}[t]
\centering
\renewcommand{\arraystretch}{1.2}
\setlength{\aboverulesep}{0pt}
\setlength{\belowrulesep}{0pt}
\resizebox{0.85\columnwidth}{!}{%
    \begin{tabular}{@{}c:c:ccccc@{}}
    \toprule
    \multirow{2}{*}{\textbf{Preds.}} & \multirow{2}{*}{\textbf{GTs}} & \multicolumn{5}{c}{\textbf{mRF1}} \\
    \cmidrule(lr){3-7}
    & & \textbf{Tiny} & \textbf{Small} & \textbf{Medium} & \textbf{Large} & \textbf{Giant} \\
    \midrule
    B & A & 20.1 & 38.2 & 49.1 & 52.1 & 72.8 \\
    C & A & 12.7 & 27.6 & 35.5 & 48.9 & 69.8 \\
    Model & A & 19.1\tiny($\pm2.8$) & 39.5\tiny($\pm3.4$) & 55.2\tiny($\pm5.8$) & 58.1\tiny($\pm4.7$) & 72.0\tiny($\pm16.1$) \\
    \hdashline
    A & B & 20.0 & 36.3 & 49.2 & 51.8 & 69.9 \\
    C & B & 16.5 & 27.3 & 40.1 & 42.9 & 59.0 \\
    Model & B & 18.9\tiny($\pm1.8$) & 33.5\tiny($\pm7.1$) & 43.6\tiny($\pm3.6$) & 43.9\tiny($\pm4.3$) & 70.1\tiny($\pm13.0$) \\
    \hdashline
    A & C & 13.0 & 27.5 & 36.8 & 50.4 & 71.5 \\
    B & C & 16.5 & 27.4 & 39.7 & 42.6 & 58.1 \\
    Model & C & 13.0\tiny($\pm1.4$) & 25.7\tiny($\pm3.0$) & 31.6\tiny($\pm6.0$) & 43.4\tiny($\pm7.6$) & 63.1\tiny($\pm12.3$) \\
    \bottomrule
    \end{tabular}%
} 
\caption{\textbf{Inter-annotator agreement on \SelvaMask's test subset.} The mRF1, stratified by crown size classes, is computed with pairs of predictions and ground truths. Expert annotators are named A, B and C, and Model is SelvaBox\fire $\rightarrow$ SAM3\fire.
}
\label{tab:detailed_per_gt_comparison}
\vspace{-10pt}
\end{table}

\subsection{Results on external datasets and OOD sites}
\label{sec:results_ood}

\paragraph{\SelvaMask enables OOD generalization.}
Table~\ref{tab:ood_results} demonstrates strong cross-biome generalization of our modular pipeline. On the tropical Detectree2 dataset, our fine-tuned approach (SelvaBox\fire $\rightarrow$ SAM3\fire) establishes a new state-of-the-art, improving mAP from 17.3 to 21.6 and mRF1 from 21.9 to 26.0.

On temperate biomes (BAMForest and QuebecTrees), both frozen and fine-tuned pipelines outperform Detectree2. The frozen pipeline (SelvaBox\ice $\rightarrow$ SAM3\ice) achieves the highest performance on BAMForest, as expected: fine-tuning on \SelvaMask specializes the model to dense, interlocking canopies, slightly reducing its transferability to temperate forests. Interestingly, fine-tuning improves QuebecTrees performance, achieving state-of-the-art global mAP by better detecting small trees and improving mRF1 by stabilizing confidence across all sizes.

For urban OAM-TCD, Detectree2 (flexi) reaches 12.3 mAP. While the frozen pipeline struggles, fine-tuning on \SelvaMask achieves 18.5 mAP, establishing a new benchmark and highlighting our dataset's cross-domain value.

\paragraph{The importance of stratified RF1 on BCI50ha.}
BCI50ha's incomplete ground truth creates misleading global metrics. The dataset primarily annotates large and giant trees, leaving smaller trees unlabeled (Fig.~\ref{fig:size_dist_all_dataset}). As a result, our model's valid predictions of tiny and small trees are wrongly counted as false positives, severely penalizing the global mRF1. However, size-stratified RF1 reveals that our method significantly outperforms all competitors on large and giant trees (Tab.~\ref{tab:ood_results}), the BCI50ha's actual focus. To support our claim, we provide qualitative results in Figure~\ref{fig:qualitative_model_predictions}. This strong stratified performance, combined with state-of-the-art tile-level mAP (37.5 vs. 29.3), confirms our model's capability despite the low global raster score.

\paragraph{Site diversity drives robustness.} We evaluate the robustness of our annotations across sites using a spatial cross-validation protocol (Sec.~\ref{sec:eval_only}).  The model generalizes remarkably well to Panama, retaining 94\% of the fully supervised performance in the zero-shot setting (17.4 vs. 18.6 mAP) (Tab.~\ref{tab:cross_site_results}). In contrast, Ecuador proves more distinct, suffering a larger drop when excluded from training (16.8 vs. 22.8 mAP). These results suggest that while \SelvaMask captures robust, transferable features across the Neotropics, including site-specific data remains beneficial for maximizing accuracy in distinct forest structures.

\begin{figure}[t]
    \centering
    \setlength{\tabcolsep}{1pt} 
    \begin{tabular}{cc}
      \includegraphics[width=0.48\linewidth]{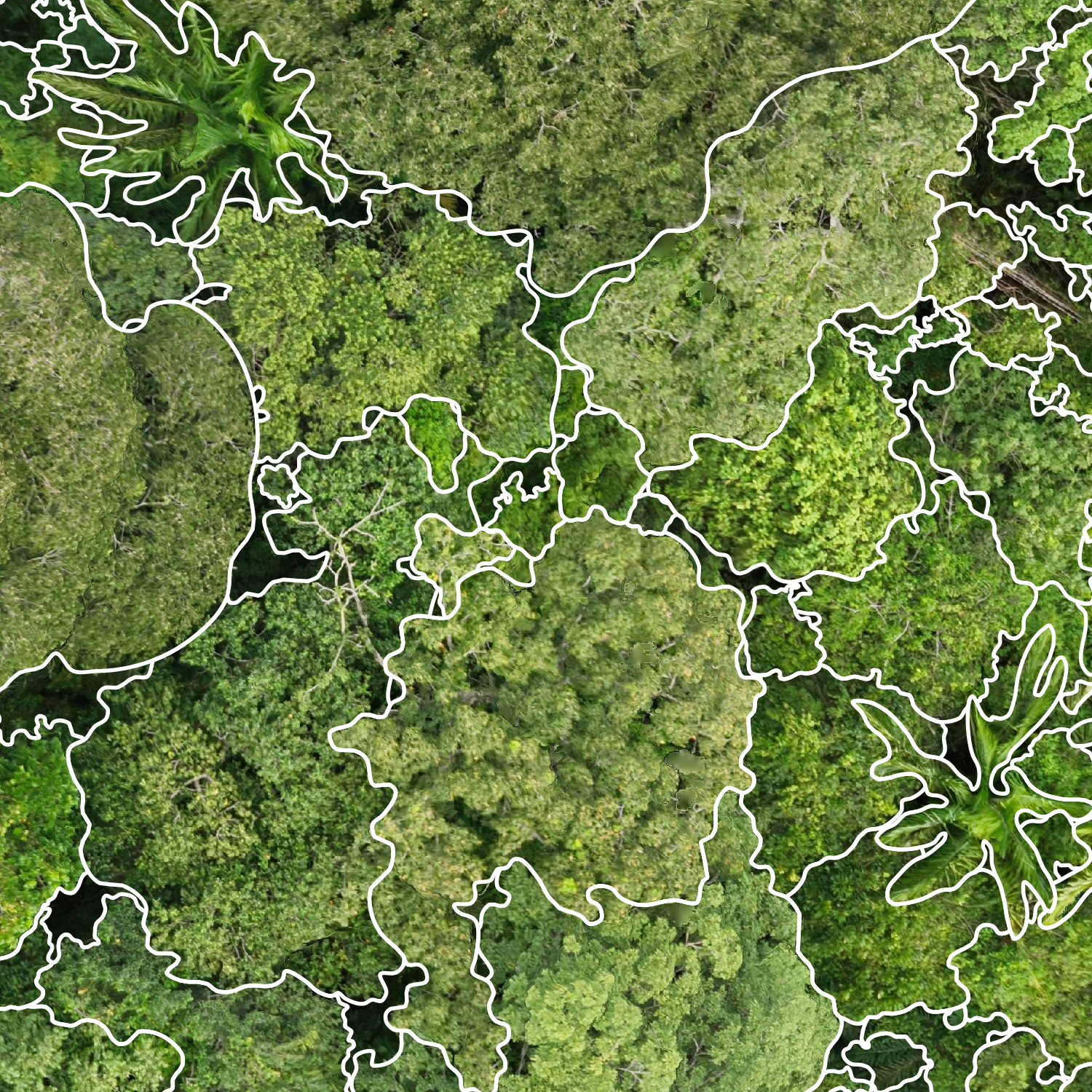} &
      \includegraphics[width=0.48\linewidth]{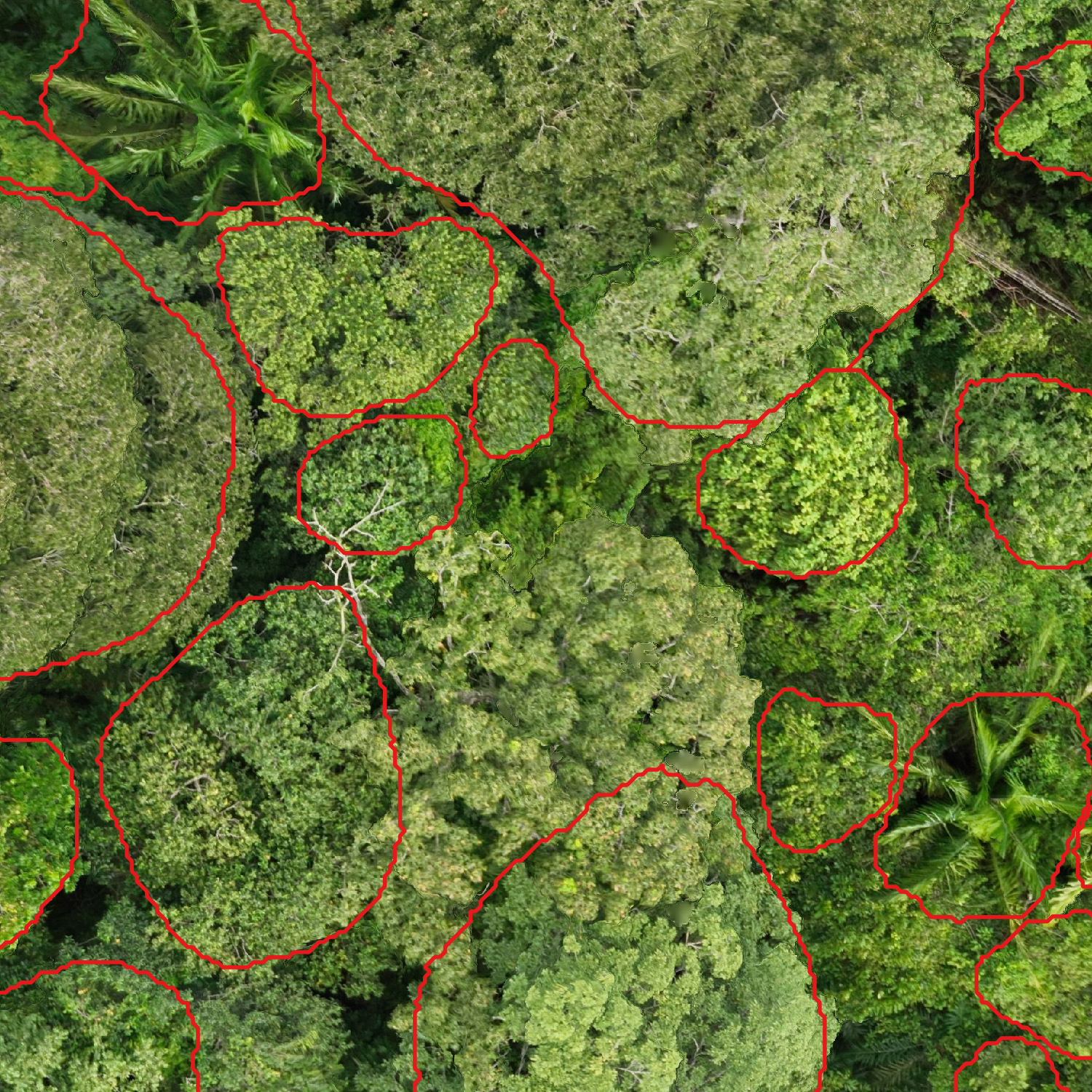} \\
      \scriptsize Ground Truth & \scriptsize Detectree2 \\[3pt] 
      
      \includegraphics[width=0.48\linewidth]{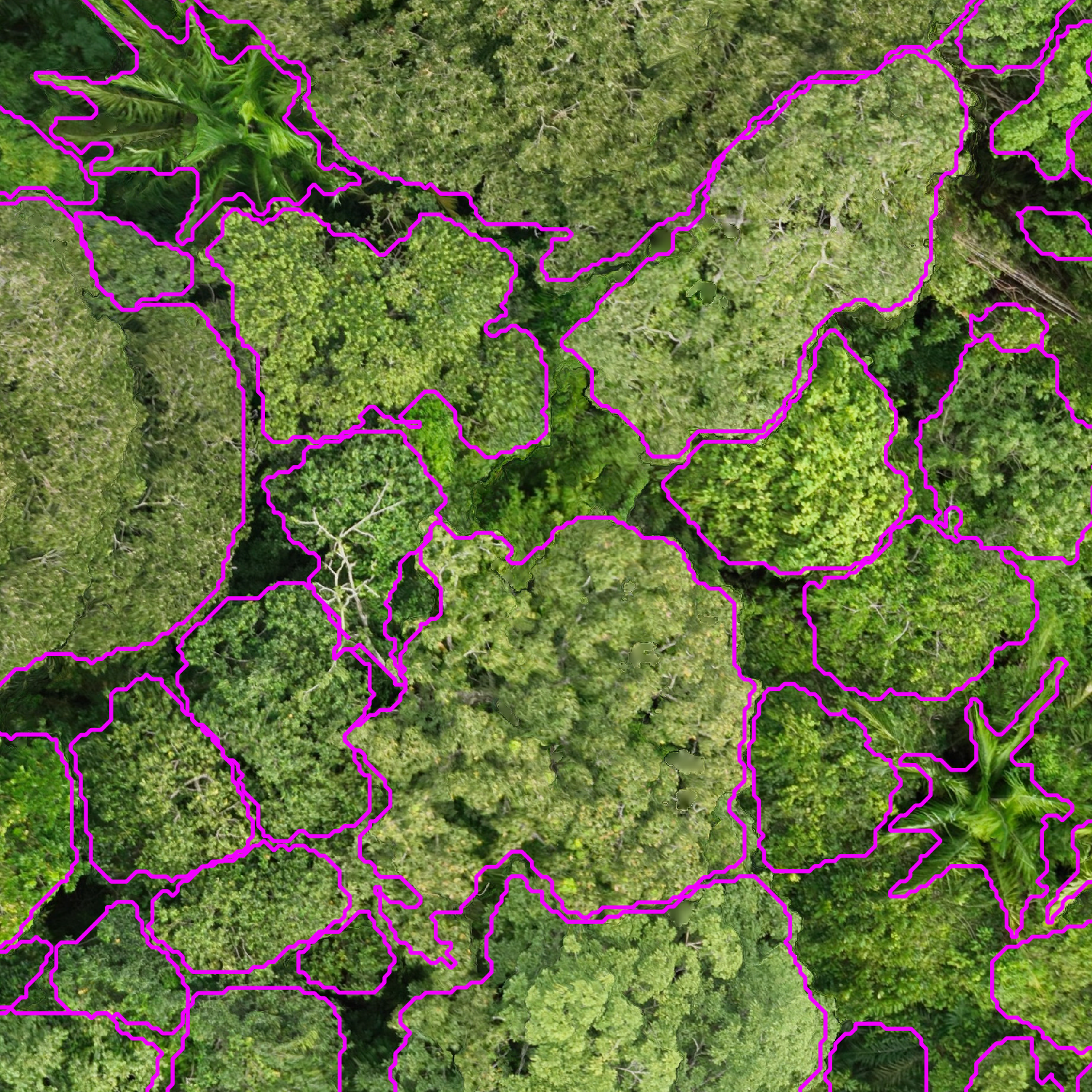} &
      \includegraphics[width=0.48\linewidth]{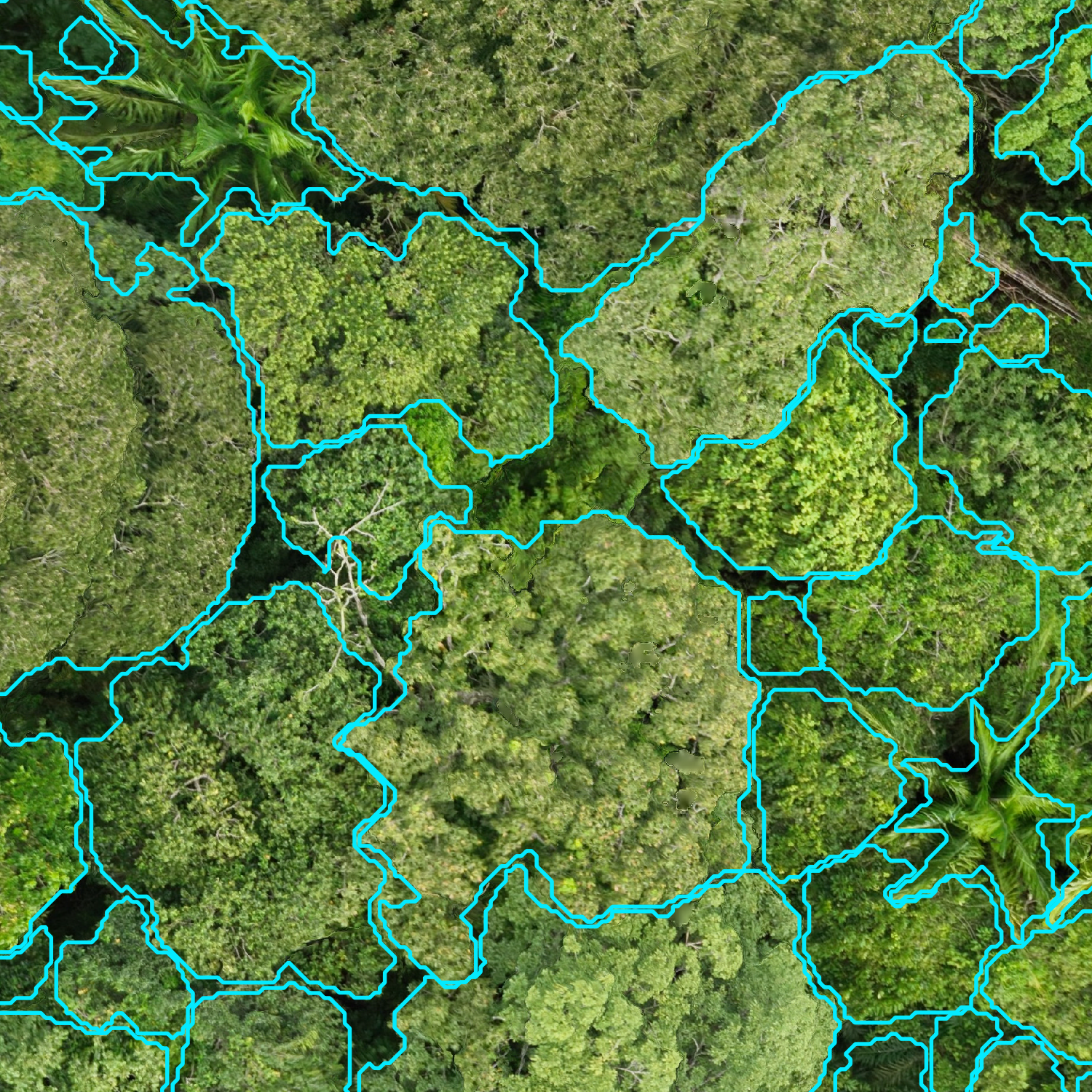} \\
      \scriptsize SelvaBox\ice $\rightarrow$ SAM3\ice & \scriptsize SelvaBox\fire $\rightarrow$ SAM3\fire
    \end{tabular}
    \caption{\textbf{Qualitative Results on \SelvaMask.} Comparison of models predictions: Detectree2 vs. our proposed modular pipeline SelvaBox$\rightarrow$ SAM3, with frozen (\ice) or fine-tuned (\fire) modules.}
    \label{fig:qualitative_model_predictions}
    \vspace{-10pt}
\end{figure}

\section{Conclusions}
In this work, we introduce \SelvaMask, the largest and most diverse tropical tree crown segmentation dataset to date, comprising 8\,861 manually delineated crowns across three Neotropical sites. By adhering to a rigorous complete-crown annotation protocol, \SelvaMask captures the extreme density of the tropical understory. Through a unified evaluation protocol combining COCO-style metrics and raster-level mRF1, we demonstrate that standard zero-shot baselines struggle in this domain, while our proposed modular pipeline, adapting VFMs via domain-specific detection-prompters, establishes a new state-of-the-art for generalized individual tree crown segmentation. This approach not only outperforms end-to-end architectures on \SelvaMask but also exhibits superior zero-shot generalization on external tropical and temperate datasets, highlighting the value of \SelvaMask for training models with strong cross-domain performance.

\paragraph{Limitations.} Despite these advances, key challenges remain. First, the sequential nature of our modular pipeline introduces a dependency bottleneck: segmentation quality is strictly upper-bounded by detection recall, meaning errors from the detector (\eg, missed small trees) propagate irreversibly to the final map. Second, our inter-annotator agreement study reveals that the task itself faces a `human ceiling'. The visual ambiguity of interlocking tropical crowns leads to substantial disagreement even among experts, suggesting that future performance gains may be difficult to quantify without multi-expert consensus ground truth. Finally, despite \SelvaMask representing three Neotropical sites, further work is needed to expand coverage of Afrotropical, Indomalayan, and Australasian regions.

\paragraph{Future Directions.} Future work should aim to bridge the gap between modular flexibility and end-to-end optimization. A promising direction is the development of fully differentiable interfaces that propagate gradients from the VFM back to the detection-prompter, allowing the modules to co-adapt rather than training in isolation. Furthermore, given the strong cross-site generalization observed in our experiments, we envision leveraging \SelvaMask alongside other emerging forest datasets to train large-scale, multi-biome foundation models capable of robust, global forest monitoring.
\section*{Acknowledgement}
We thank Marie-Jeanne Gascon-DeCelles, Sabrina Demers-Thibeault, and Lazare Magué for their work on data annotation. We are also grateful to Vincent Le Falher and Antoine Caron-Guay for their support with the laboratory infrastructure. This research was enabled in part by computing resources provided by Mila (Quebec AI Institute) and the Digital Research Alliance of Canada. This project was undertaken thanks to funding from IVADO, including the PRF3 project 'AI, biodiversity, and Climate Change', a Canada Research Chair and a Discovery Grant from NSERC to E.L., the Ceiba Fund from the Population Biology Foundation, and the Angèle St-Pierre \& Hugo Larochelle Research Chair in AI for the Environment.

\section*{Impact Statement}
Our research bridges the gap between state-of-the-art computer vision and ecological monitoring in complex tropical environments. By successfully adapting Vision Foundation Models for dense canopy segmentation, we demonstrate that generalist AI can be effectively specialized for high-impact scientific domains. This work offers a practical, low-cost solution for automated forest census, enabling the precise estimation of carbon stocks and biodiversity needed to support conservation efforts and climate change mitigation. However, we acknowledge that high-precision canopy mapping carries a potential risk of misuse, such as the targeting of high-value timber for illegal extraction. We emphasize that this work is intended strictly to empower conservationists and local stakeholders in preserving threatened ecosystems.
\FloatBarrier
\bibliography{references}
\bibliographystyle{icml2026}

\newpage
\appendix
\onecolumn
\raggedbottom

\section{SelvaMask dataset}

\subsection{Image acquisition metadata}
\label{appendix:Image_Acquisition}
Table~\ref{tab:flight_metadata} details the specific flight parameters and acquisition conditions for the three \textsc{SelvaMask} sites. These parameters were chosen to maximize orthomosaic quality and minimize motion blur (shutter speed $1/1000$s) across diverse lighting conditions.

\begin{table}[h]
\centering
\small
\renewcommand{\arraystretch}{1.2}
\setlength{\tabcolsep}{6pt} 
\begin{tabular}{@{}l c c c@{}}
\toprule
\textbf{Parameter} & \textbf{Panama (BCI)} & \textbf{Brazil (ZF2)} & \textbf{Ecuador (Tiputini)} \\
\midrule
\multicolumn{4}{l}{\textit{Flight Parameters}} \\
\textbf{Date} & Nov 22, 2024 & Jan 31, 2024 & Jun 13, 2024 \\
\textbf{Time (Start--End)} & 11:05 -- 11:39 & 14:47 -- 15:06 & 10:57 -- 11:27 \\
\textbf{Platform} & DJI Mavic 3M & DJI Mavic 3M & DJI Mavic 3E \\
\textbf{Sensor Type} & 20MP 4/3 CMOS & 20MP 4/3 CMOS & 20MP 4/3 CMOS \\
\textbf{Altitude (AGL)} & 60 m & 43 m & $\sim$90 m$^{\ast}$ \\
\textbf{Overlap (Front/Side)} & 85\% / 75\% & 85\% / 75\% & 85\% / 70\% \\
\textbf{Shutter Speed} & $1/1000$ s & $1/1000$ s & $1/1000$ s \\
\textbf{Scanning Mode} & Repetitive & Repetitive & Repetitive \\
\textbf{Conditions} & Partly cloudy skies & Clear skies & Clear skies \\
\midrule
\multicolumn{4}{l}{\textit{Orthomosaic Properties}} \\
\textbf{Dimensions (px)} & $48\,712 \times 41\,463$ & $22\,947 \times 21\,047$ & $31\,503 \times 41\,838$ \\
\textbf{Resolution (GSD)} & 1.83 cm/px & 1.34 cm/px & 3.47 cm/px \\
\textbf{Covered Area} & 67.8 ha & 8.7 ha & 158.3 ha \\
\bottomrule
\end{tabular}
\caption{Detailed acquisition metadata and resulting orthomosaic properties for all sites. Altitude is reported Above Ground Level (AGL). Time is local. $^{\ast}$Altitude estimated based on GSD.}
\label{tab:flight_metadata}
\end{table}
\subsection{Annotation protocol}
\label{app:annotation_guidelines}

\paragraph{Software and setup.} All annotations were performed using ArcGIS Pro. Annotators were provided with high-resolution RGB orthomosaics and a predefined Area of Interest (AOI) layer to ensure consistent spatial coverage. 

\begin{enumerate}
    \item \textbf{Initialization:} Load the site-specific orthomosaic and the Area of Interest (AOI) mask. Create a new polygon feature class for the session.
    \item \textbf{Tracing Method:} Use the \textit{Freehand Autocomplete} tool to trace crown boundaries. 
    \begin{itemize}
        \item For isolated trees: Trace the full contour.
        \item For clustered trees: Start the trace inside an existing neighbor's polygon, trace the new crown's outer edge, and end the trace back inside the neighbor's polygon. The software automatically calculates the shared boundary to ensure zero overlap.
    \end{itemize}
    \item \textbf{Inclusion Criteria:} 
    \begin{itemize}
        \item All visible canopy trees must be annotated.
        \item Trees straddling the edge of the AOI must be included.
    \end{itemize}
    \item \textbf{Exclusion Criteria:} 
    \begin{itemize}
        \item Do not annotate shadows, deep understory, or ambiguous blurry regions.
        \item Delete any ``sliver'' or ``gap'' polygons accidentally created by the autocomplete tool in shadowed regions.
    \end{itemize}
\end{enumerate}

\clearpage
\FloatBarrier
\subsection{Dataset Statistics}
\label{appendix:dataset_statistics}

\begin{figure}[H]
    \centering
    \includegraphics[width=0.75\linewidth]{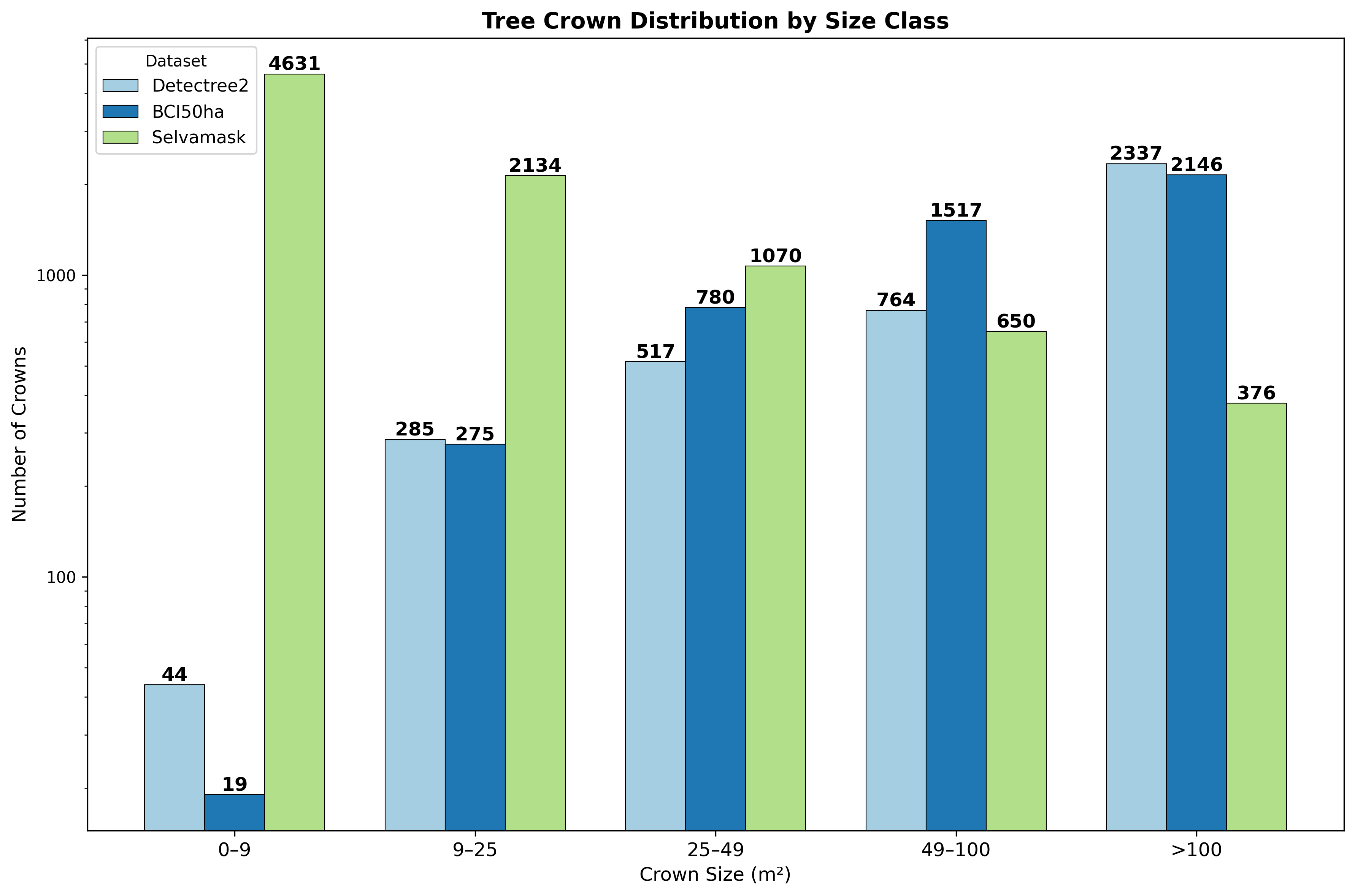}
    \caption{Crown area distribution in for all tropical datasets in this study. Vertical lines indicate the five ecological size classes (Tiny, Small, Medium, Large, Giant). Y-axis is in logarithmic scale.}
    \label{fig:size_dist_all_dataset}
\end{figure}

\FloatBarrier
\subsection{Tiling strategy}
\label{appendix:split_zones}

\begin{figure}[H]
    \centering
    \includegraphics[width=0.75\linewidth]{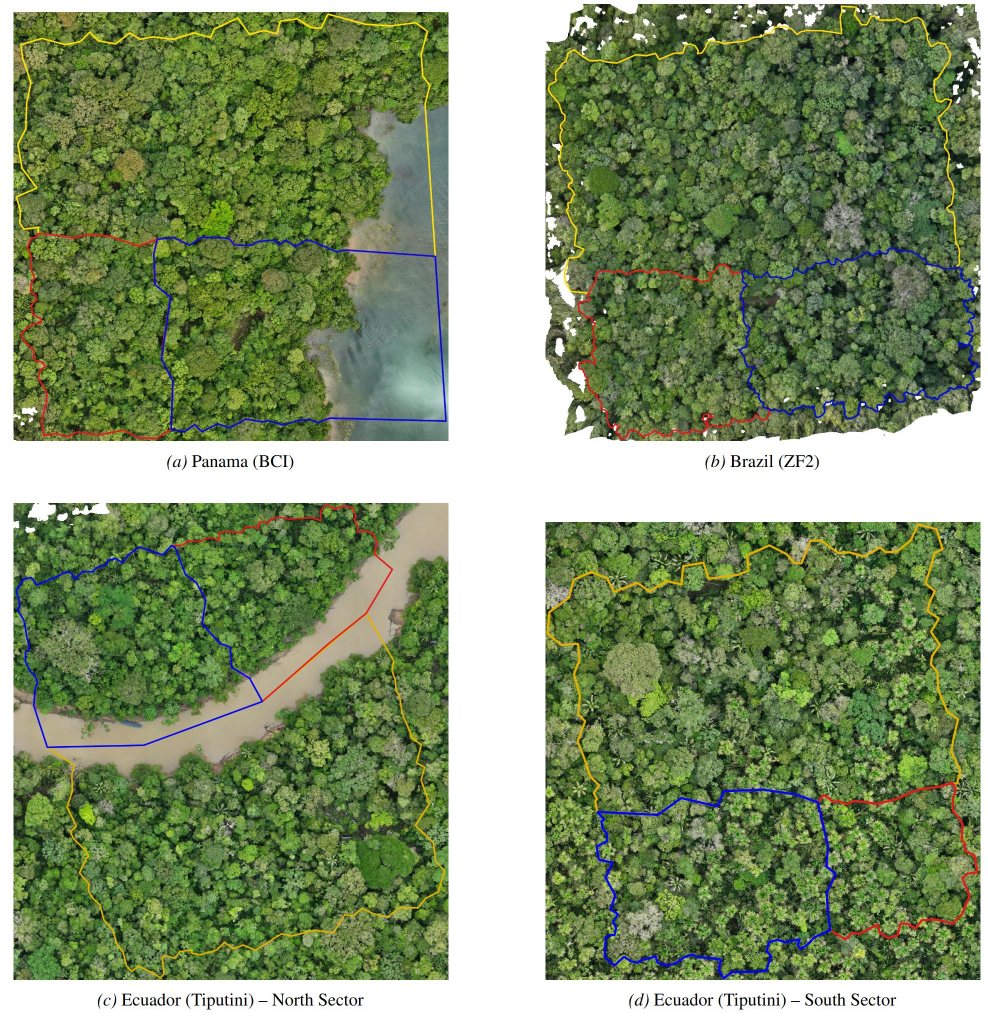}
    \caption{\textbf{Spatial Cross-Validation Splits.} Visualization of the Training (Yellow), Validation (Red), and Test (Blue) zones across the three SELVAMASK sites. The Ecuador site is split into two nearby sectors (c, d) within the same orthomosaic. Pixels outside these zones are masked to ensure strict spatial isolation.}
    \label{fig:selvamask_splits_zones}
\end{figure}

\clearpage
\section{Training parameters}
\label{appendix:training}

\subsection{Data augmentation}
\label{appendix:data_augmentation}
Listed in Table \ref{tab:augmentation_params} are the data augmentation parameters used for all segmentors and detection models. Note that these correspond to the training augmentations used in \cite{baudchon_selvabox_2026} for their multi-resolution, multi-datasets experiments.

\begin{table}[h]
\centering
\small

\begin{tabular}{@{}ll@{}}
\toprule
\textbf{Parameter} & \textbf{Value} \\
\midrule
\multicolumn{2}{l}{\textit{Color \& Intensity}} \\
Brightness & $0.2$ (Prob: $0.5$) \\
Contrast & $0.2$ (Prob: $0.5$) \\
Saturation & $0.2$ (Prob: $0.5$) \\
Hue & $10$ (Prob: $0.3$) \\
\midrule
\multicolumn{2}{l}{\textit{Geometric Transformations}} \\
Horizontal Flip & True \\
Vertical Flip & True \\
Rotation & $\pm 30^{\circ}$ (Prob: $0.5$) \\
\midrule
\multicolumn{2}{l}{\textit{Cropping \& Resizing}} \\
Crop Probability & $0.5$ \\
Crop Min Intersection & $0.5$ \\
Train Crop Size Range & $[666, 2666]$ pixels \\
Target Image Size & $\{1024, 1777\}$ pixels \\
Early Conditional Size & $2000$ pixels \\
\bottomrule
\end{tabular}
\caption{Data augmentation hyperparameters used during training. These transformations are applied on-the-fly to improve model robustness to lighting and orientation changes.}
\label{tab:augmentation_params}
\end{table}

\subsection{NMS parameters optimization for mRF1 metric\label{appendix:nms_params}}
The mRF1 metric requires optimizing NMS and confidence thresholds on
a validation set prior to benchmarking on the test set \cite{baudchon_selvabox_2026}.
In this work, we optimize these hyperparameters for Detectree2 \cite{ballAccurateDelineationIndividual2023} on the original Detectree2 dataset, and for all other methods on \SelvaMask. 
This ensures that Detectree2, our primary baseline for tropical tree crown segmentation, remains strictly OOD relative to \SelvaMask during evaluation.
The only exception is for the QuebecTrees dataset in our OOD evaluation (Tab.~\ref{tab:ood_results}), where the NMS is optimized on the validation set of QuebecTrees itself for all methods due to the significant domain shift. For the rest of the methodology, i.e the grid search setup, we follow \cite{baudchon_selvabox_2026}.

\subsection{SAM fine-tuning}
\subsubsection{Hyperparameters}
\label{appendix:Hyperparameters}

\begin{table}[H]
\centering
\small

\begin{tabular}{@{}ll@{}}
\toprule
\textbf{Hyperparameter} & \textbf{Value} \\
\midrule
\multicolumn{2}{l}{\textit{Models}} \\
SAM2 Variant & facebook/sam2-hiera-large \\
SAM3 Variant & SAM3-Large \\
\midrule
\multicolumn{2}{l}{\textit{Optimization}} \\
Batch Size & 2 \\
Max Prompts per Batch & 64 \\
Total Training Steps & 4,300 \\
Optimizer & AdamW \\
Scheduler & Cosine Decay \\
Warmup & 5\% Linear Warmup \\
LR (Image Encoder) & $1.0 \times 10^{-5}$ \\
LR (Prompt Enc / Mask Dec) & $3.0 \times 10^{-5}$ \\
Weight Decay & $0.01$ \\
\midrule
\multicolumn{2}{l}{\textit{Prompting Strategy}} \\
Prompt Source & Detector-predicted boxes \\
Box Jitter (Noise) & 0.0 \\
Point Prompts & False \\
\midrule
\multicolumn{2}{l}{\textit{Loss Function}} \\
Loss Components & Dice + IoU \\
$\lambda_{\text{dice}}$ & $1.0$ \\
$\lambda_{\text{iou}}$ & $1.0$ \\
$\lambda_{\text{focal}}$ & $0.0$ (Disabled) \\
IoU Regression Loss & L1 \\
\bottomrule
\end{tabular}
\caption{\textbf{SAM fine-tuning Hyperparameters.} Common configuration used for both SAM2 and SAM3 experiments. We fine-tuned end-to-end using Automatic Mixed Precision (AMP) with detector-predicted box prompts.}
\label{tab:sam_hyperparameters}
\end{table}

\subsubsection{SAM training components}\label{appendix:SAM_components}

\begin{table}[h!]
\centering
\small
\setlength{\tabcolsep}{8pt}
\renewcommand{\arraystretch}{1.2}
\begin{tabular}{@{}ccc|ccc@{}}
\toprule
\multicolumn{3}{c}{\textbf{Component}} & \multicolumn{3}{c}{\textbf{Performance}} \\
\cmidrule(r){1-3} \cmidrule(l){4-6}
\textbf{Image Enc} & \textbf{Prompt Enc} & \textbf{Mask Dec} & \textbf{mAP} & \textbf{mAR} & \textbf{mRF1} \\
\midrule
\ice & \ice & \ice & 16.5 & 30.6 & 29.8 \\ 
\ice & \ice & \fire & 16.7 & 29.0 & 28.1 \\ 
\ice & \fire & \ice & 16.5 & 30.6 & 29.8 \\
\ice & \fire & \fire & 16.7 & 29.0 & 28.2 \\
\fire & \ice & \ice & 18.1 & 31.3 & 29.9 \\
\fire & \ice & \fire & \underline{18.3} & \textbf{32.6} & \textbf{31.2} \\
\fire & \fire & \ice & 17.1 & 30.9 & 29.3 \\
\fire & \fire & \fire & \textbf{19.5} & \underline{31.5} & \underline{31.1} \\ 
\bottomrule
\end{tabular}
\caption{\textbf{Ablation Study on SAM2 fine-tuning Strategy.} We evaluate the impact of freezing (\ice) vs. fine-tuning (\fire) different components of the SAM2 architecture. We mark the best and second-best scores in \textbf{bold} and \underline{underline}, respectively. All experiments were run for 1500 steps, with the same Hyperparameters mentioned in Table~\ref{tab:sam_hyperparameters}}
\label{tab:ablation_sam_components}
\end{table}

\clearpage
\subsection{End-to-end model training hyperparameters}

\begin{table*}[h]
\small
\centering
\renewcommand{\arraystretch}{1.2}
\resizebox{0.8\textwidth}{!}{
    \begin{tabular}{@{}lccccc@{}}
    \toprule
    \multirow{2}{*}{\textbf{Hyperparameter}} & \textbf{Mask R-CNN} & \multicolumn{2}{c}{\textbf{Mask2Former}} & \multicolumn{2}{c}{\textbf{RSPrompter (Anchor)}} \\
    \cmidrule(lr){2-2} \cmidrule(lr){3-4} \cmidrule(lr){5-6}
     & ResNet50 & ResNet50 & Swin-L & SAM-Base & SAM-Huge \\
    \midrule
    \multicolumn{6}{l}{\textit{Architecture}} \\
    Backbone Pretraining & ImageNet-1k & ImageNet-1k & ImageNet-22k & SA-1B & SA-1B \\
    Model Pretraining & COCO & COCO & COCO & SA-1B & SA-1B \\
    \midrule
    \multicolumn{6}{l}{\textit{Optimization}} \\
    Optimizer & SGD & AdamW & AdamW & AdamW & AdamW \\
    Base Learning Rate & $5.0 \times 10^{-3}$ & $1.0 \times 10^{-4}$ & $1.0 \times 10^{-4}$ & $1.0 \times 10^{-5}$ & $1.0 \times 10^{-5}$ \\
    Batch Size & 4 & 4 & 4 & 4 & 4 \\
    Max Epochs & 800 & 800 & 800 & $\approx1200$ & $\approx1200$ \\
    Box Prompts & - & - & - & 64 & 64 \\
    \midrule
    \multicolumn{6}{l}{\textit{Scheduler}} \\
    Type & WarmupCosine & WarmupCosine & WarmupMultiStep & WarmupCosine & WarmupCosine \\
    Decay Milestones & - & - & [80\%, 90\%] & - & - \\
    Warmup Iterations & 100 & 100 & 100 & 50 & 50 \\
    \bottomrule
    \end{tabular}
}
\caption{\textbf{Hyperparameter configurations for all trained models.} We detail the optimization settings for the baselines (Mask R-CNN, Mask2Former) and our prompt-based methods (RSPrompter variants). We report only the hyperparameters modified for our experiments; all others follow the default configurations of their respective codebases.}
\label{tab:all_hyperparams}
\end{table*}

\FloatBarrier
\clearpage
\section{Additional results\label{appendix:additional_results}}

\begin{figure}[!h]
    \centering
    \includegraphics[width=\linewidth]{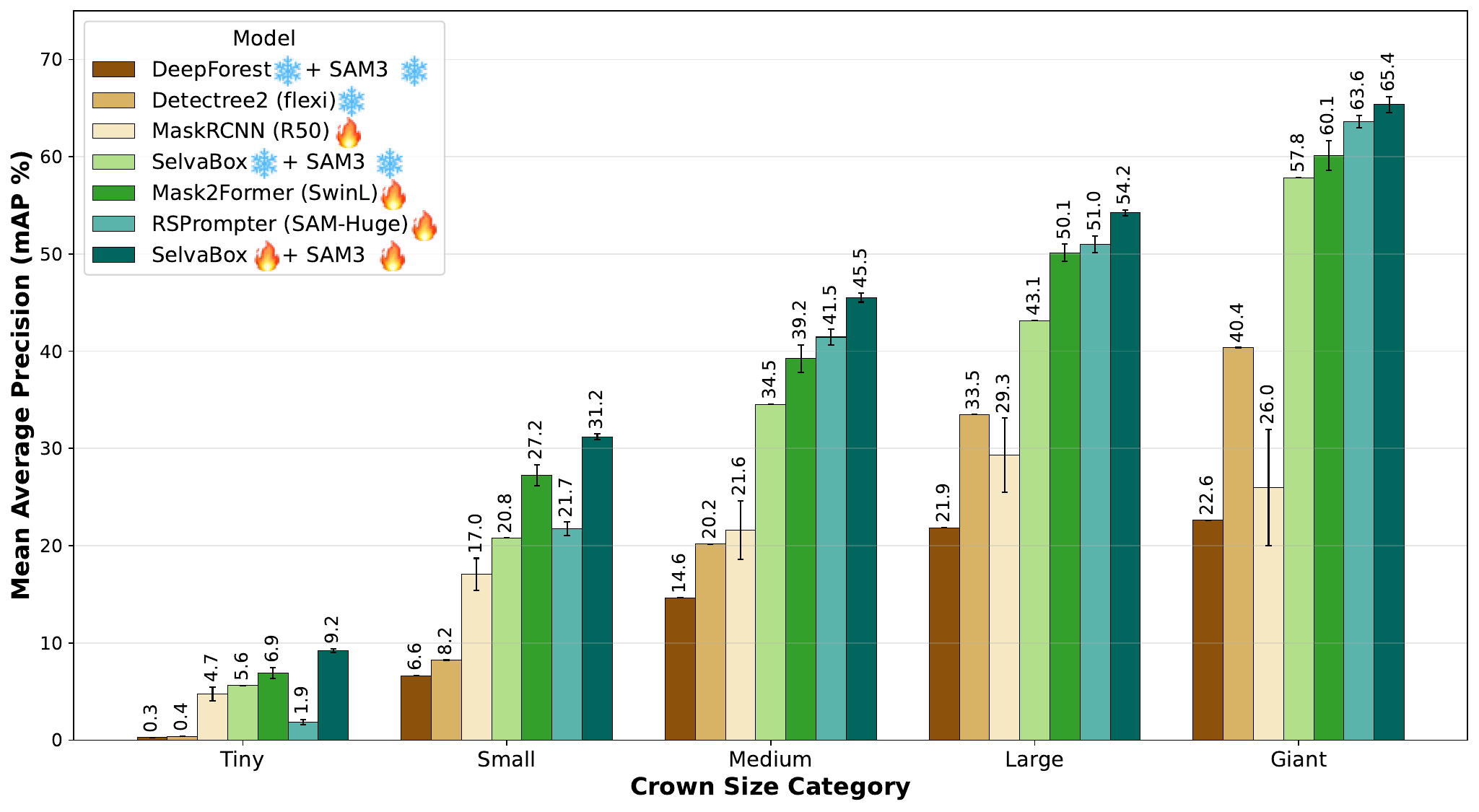}
    \caption{Different model performances in mAP by size category (Tiny, Small, Medium, Large, Giant).}
    \label{fig:map_by_size}
\end{figure}

\begin{figure}[!h]
    \centering
    \includegraphics[width=\linewidth]{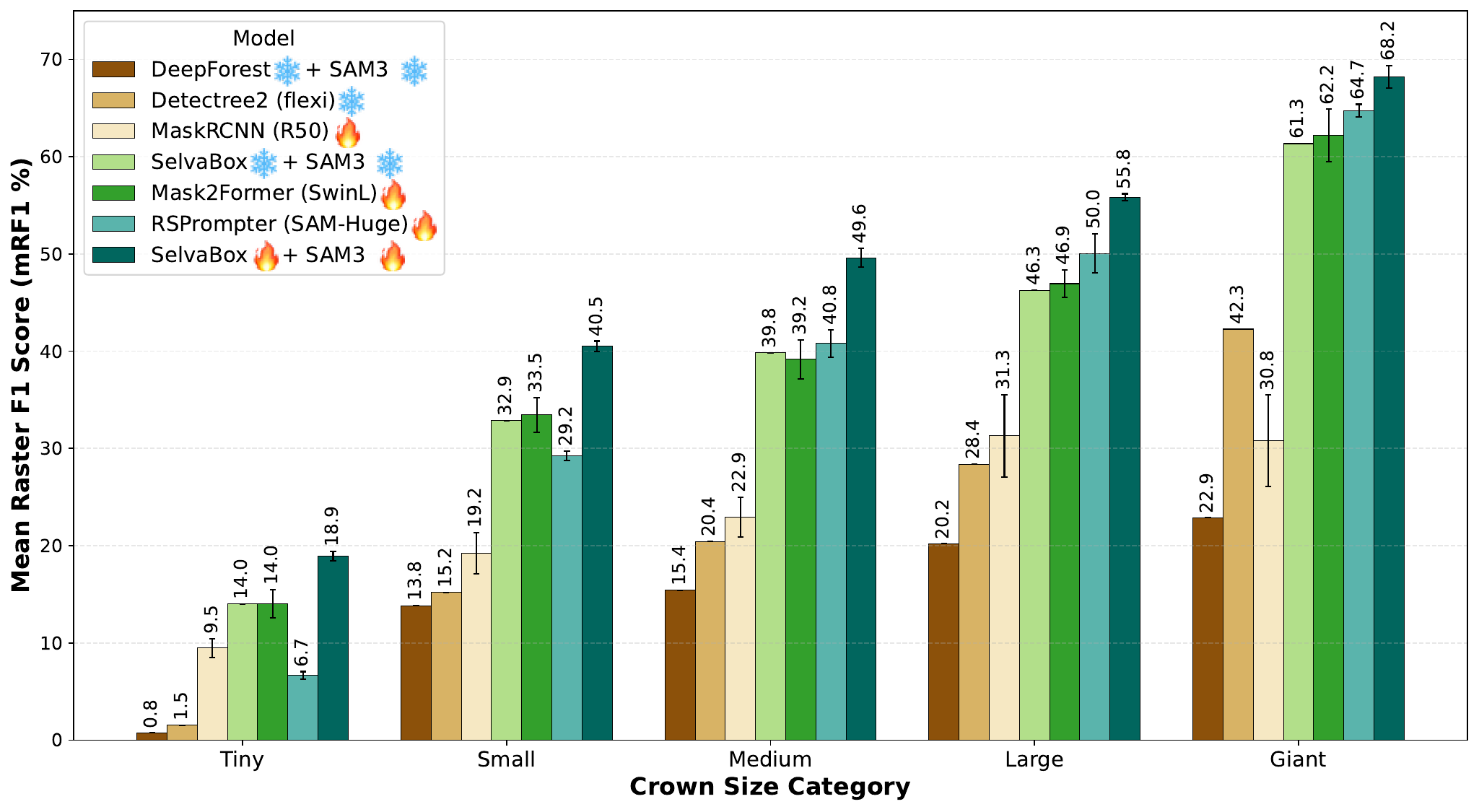}
    \caption{Different model performances in mRF1 by size category (Tiny, Small, Medium, Large, Giant).}
    \label{fig:mrf1_by_size}
\end{figure}

\begin{table*}[!h]
\centering
\scriptsize
\setlength{\tabcolsep}{2.8pt}
\renewcommand{\arraystretch}{1.2}
\setlength{\aboverulesep}{0pt}
\setlength{\belowrulesep}{0pt}
\begin{tabular}{@{}l:ccc:ccc:ccc@{}}
\toprule
\textbf{Method} & $\Delta$mAP & $\Delta$AP$_{50}$ & $\Delta$AP$_{75}$ & $\Delta$mAR & $\Delta$AR$_{50}$ & $\Delta$AR$_{75}$ & $\Delta$mRF1 & $\Delta$RF1$_{50}$ & $\Delta$RF1$_{75}$ \\
\midrule
SAM2\fire\ - SAM2\ice & +2.7 & +4.0 & +3.8 & +2.1 & +0.3 & +3.3 & +3.2 & +1.8 & +4.4 \\
SAM3\fire\ - SAM3\ice & +2.4 & +1.0 & +4.0 & +2.2 & +0.0 & +4.3 & +3.5 & +1.9 & +5.2 \\
\bottomrule
\end{tabular}
\caption{Effect of fine-tuning on SelvaBox$\rightarrow$SAM (absolute $\Delta$ in percentage points over the frozen / zero-shot variant)}
\label{tab:ft_gains_pp}
\end{table*}

\begin{figure}[t]
    \centering
    \includegraphics[width=0.75\linewidth]{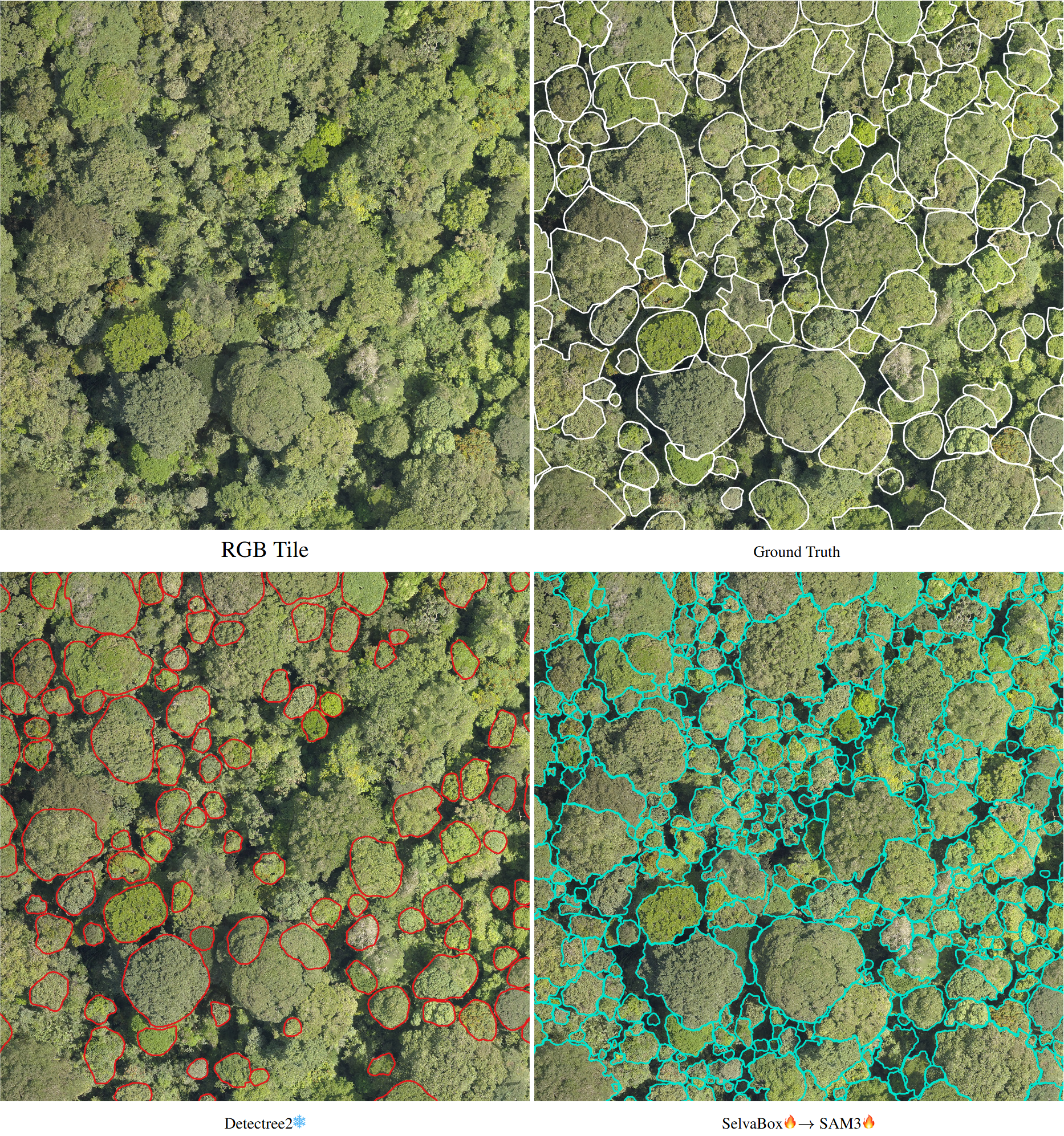}
    \caption{\textbf{Qualitative Results on BCI50ha.} Comparison of models predictions: Detectree2 vs. our proposed modular pipeline SelvaBox$\rightarrow$ SAM3 with fine-tuned (\fire) modules. Note that the BCI50ha dataset annotation protocol omits small understory trees. Similarly, Detectree2 produces sparse predictions. In contrast, our pipeline delineates all visible crowns, successfully recovering the dense small trees that are ignored by both the baseline and the dataset annotations.}
    \label{fig:qualitative_model_predictions}
\end{figure}
%
%

\end{document}